\documentclass{amsart}

\usepackage{url}
\usepackage{amsmath}
\usepackage{amsfonts}
\usepackage{amssymb}
\usepackage{mathtools}
\usepackage{xcolor}
\usepackage{graphicx}
\usepackage{multirow}
\usepackage{algpseudocode, algorithm}

\DeclareMathOperator{\XX}{\boldsymbol{X}}
\DeclareMathOperator{\SX}{\boldsymbol{S}}
\DeclareMathOperator{\ZX}{\boldsymbol{Z}}

\newtheorem{definition}{Definition}

\newtheorem{Assumption}{Assumption}

\begin{document}

\title[]{Design a Metric Robust to Complicated High Dimensional Noise for Efficient Manifold Denoising}

\author{Hau-Tieng Wu}
\address{Courant Institute of Mathematical Sciences, New York University, New York, NY, USA}
\email{hauwu@cims.nyu.edu}

\maketitle

\begin{abstract}
In this paper, we propose an efficient manifold denoiser based on landmark diffusion and optimal shrinkage under the complicated high dimensional noise and compact manifold setup. It is flexible to handle several setups, including the high ambient space dimension with a manifold embedding that occupies a subspace of high or low dimensions, and the noise could be colored and dependent. A systematic comparison with other existing algorithms on both simulated and real datasets is provided. This manuscript is mainly algorithmic and we report several existing tools and numerical results. Theoretical guarantees and more comparisons will be reported in the official paper of this manuscript. 

\end{abstract}

\section{Introduction}

Matrix is probably the most common data structure researchers encounter in practice. Usually, researchers have a data matrix $\XX\in \mathbb{R}^{p\times n}$ at hand, which saves $n$ points in $\mathbb{R}^p$; that is, we have a point cloud $\mathcal{X}:=\{x_i\}_{i=1}^n\subset \mathbb{R}^p$, where $x_i$ is the $i$-th column of $\XX$. Due to the inevitable noise, $\XX$ is modeled as $\XX=\SX+\Xi$, where $\SX$ is the {\em clean} data and $\Xi$ contains inevitable noise so that each column of $\ZX$ has mean $0$ and finite variance. Denote $\mathcal{S}:=\{s_i\}_{i=1}^n$ be the clean point cloud. Moreover, $p$ is usually high. The data analysis goal is extracting usual information from  the high dimensional  dataset $\XX$, and we desire that the information extraction process is robust to the existence of $\Xi$, in the sense that the extracted information from $\XX$ is ``close'' to that extracted from the clean data $\SX$. One specific example is recovering $\SX$ from $\XX$, which could be understood as the {\em denoising} mission, which is a specific example of the general ``data sharpening'' mission \cite{choi2000data}.

To proceed with the denoising mission, we put assumptions on $\SX$ and $\Xi$. In this paper, we focus on the {\em low dimensional manifold} model for the clean dataset $\SX$; that is, the point cloud $\mathcal{X}$ is supported on a low dimensional Riemannian manifold $(M,g)$ isometrically embedded in $\mathbb{R}^p$ via $\iota$. Moreover, we assume the {\em high dimensional setup}; that is, we assume that $p$ is high in the sense that $p=p(n)\to \infty$ as $n\to \infty$. For the noise $\Xi$, to better capture the real world complicated situation, we assume that $\Xi$ satisfies the {\em separable covariance structure}; that is, $\Xi = A^{1/2}ZB^{1/2}$,
where the entries of $Z$ are i.i.d. with zero mean and mild moment conditions that will be specified later, and $A$ and $B$ are $p \times p$ and $n \times n$ deterministic positive-definite matrices respectively. Here, $A$ describes the colorness of noise, and $B$ describes how the noise is dependent across samples. We call this setup the {\em low dimensional manifold with high dimensional dependent noise model}. This low dimensional manifold with high dimensional dependent noise model is motivated by biomedical time series, or more specifically the {\em single channel blind source separation} problem. See Section \ref{section: LFP DBS example} below for an example.

There are roughly two common approaches toward analyzing the noisy dataset $\mathcal X$ under the above setup. 
First, design and study a robust almost isometric embedding algorithm so that the noise does not impact too much and it can be directly applied to the noisy data. With theoretical guarantee, researchers can view the embedded point clouds as ``clean dataset'' and proceed with the analysis. There have been several works in this direction \cite{ElKaroui:2010a,Steinerburger2016,el2016graph,shen2022robust,shen2022scalability,9927456}. In \cite{Steinerburger2016}, the robustness of DM is explored under the assumption that the neighbor information is available. It is shown in \cite{ElKaroui:2010a,el2016graph} that when the neighbor information is missing, DM is still robust to noise, and if it is slightly perturbed by design (see below), it is robust to large noise. In \cite{9927456}, this robustness property is explored from the RMT perspective. The robustness in this approach is usually quantified in the $L^2$ norm sense, and the desired properties are explored directly from the noisy dataset. In \cite{shen2022robust,shen2022scalability}, a modification of DM called {\em RObust and Scalable Embedding via LANdmark Diffusion} (ROSELAND) is introduced. It is shown to enjoy properties of DM and robust to noise.

Second, denoise the manifold first before studying desired properties. It is important to clarify and distinguish the mission of {\em manifold denoising} and a closely related mission of {\em manifold fitting}. In the manifold denoising problem, we aim to denoise each sample from the noisy dataset and recover the associated clean sample {\em on the manifold}, while in the manifold fitting problem, we aim to recover the manifold from the (possibly) noisy dataset. Intuitively, we can denoise the manifold first and fit the manifold by some ``interpolation'' idea, or we can fit the manifold first and denoise the dataset by projecting the noisy dataset to the clean manifold. In this sense they are closely related.
The manifold fitting problem can be traced back to the work on principle curve and surface \cite{hastie1989principal} and recent development in the principle manifold framework \cite{meng2021principal}. Since the fitting approach in this direction needs a modification for the manifold denosing purpose and it is out of the scope of this paper, we skip them here and refer readers with interest to \cite{JMLR:v12:ozertem11a,meng2021principal} for a nice review of various variations. After \cite{hastie1989principal}, several manifold fitting algorithms were developed, including fitting a locally linear set to the noisy manifold dataset \cite{gong2010locally}, ridge estimation based on probability density function \cite{JMLR:v12:ozertem11a,genovese2014nonparametric,mohammed2017manifold}, Fitting Putative Manifold (FPM) by local principal component analysis  \cite{pmlr-v75-fefferman18a,fefferman2019fitting}, Manifold reconstruction via Gaussian processes (MrGap) \cite{dunson2022inferring}, Manifold Locally Optimal Projection (MLOP) \cite{faigenbaum2023manifold} and a modified FPM (mFPM) \cite{yao2023manifold}\footnote{In \cite{pmlr-v75-fefferman18a,fefferman2019fitting,yao2023manifold}, the authors do not name their algorithms, and we use the title in \cite{pmlr-v75-fefferman18a} as the algorithm's name here.} In a recent paper, the manifold fitting for dataset with large noise is considered \cite{fefferman2023fitting} when the manifold satisfies the ``R-exposedness'' property.
These manifold fitting algorithms are suitable for the manifold denoising except MLOP \cite{faigenbaum2023manifold}. MLOP views the noisy dataset as a guidance to deform the noisy dataset following some rules, including repulsive rules inside the deformed dataset and the similarity of the deformed dataset and the noisy dataset, which leads to a deformed dataset distributed on the manifold as even as possible but does not denoise each sample point. 
The most common approach to denoise the manifold is applying the truncated singular value decomposition (TSVD) to ``denoise the matrix'', or reduce the impact of noise \cite{golub1965calculating}, under the low rank assumption.
Unfortunately, it does not work when the rank of the manifold support is high. For example, consider a $1$-dim smooth manifold without boundary parametrized by 
\begin{align}
\iota_p:\theta\in &[0,2\pi)\to [\alpha_1\cos(\theta),\alpha_2\sin(\theta),\\
&\alpha_3\cos(2\theta),\alpha_4\sin(2\theta),\ldots,\alpha_{2p-1}\cos(p\theta),\alpha_{2p}\sin(p\theta)]^\top \in \mathbb{R}^{2p}\,,\nonumber
\end{align}
where $\alpha_l>0$ for all $l=1,\ldots,2p$.
Clearly, $n$ uniformly independently sampled $\theta_i$, $i=1,\ldots,n$, lead to $n$ points $x_i:=\iota_p(\theta_i)$ in $\mathbb{R}^{2p}$, and the associated data matrix is of rank $2p$. Note that this is not against Nash's isometric embedding theorem \cite{nash1956imbedding}, which states the existence of an isometric embedding into a low dimensional Euclidean space. But in practice the data generation mechanism may not follow the Nash's isometric embedding, and hence TSVD fails in this case.
The nonlinear approach started from the pioneering inverse diffusion work \cite{NIPS2006_2997}. After that, several modern tools taking local geometric structure into account into account, including nonlocal mean or median \cite{su2019OSonEEG,alagapan2019diffusion}, denoising by Nolinear Robust Principal Analysis (NRPCA) \cite{NEURIPS2019_a76c0abe} and Manifold Moving Least-Squares (MMLS) \cite{sober2020manifold,aizenbud2021nonparametric}. 
Those manifold denoising algorithms can be roughly classified into two categories.
The first one is fitting a manifold to the noisy dataset by finding the ridge of some preassigned functions. For example, the kernel density estimation (KDE) is used in \cite{genovese2014nonparametric}, the ``approximate square distance function'' \cite{fefferman2016testing} constructed from KDE or tangent space estimation is used in \cite{mohammed2017manifold}, 
and the distance function to the manifold is used in \cite{pmlr-v75-fefferman18a,fefferman2019fitting,yao2023manifold}. Note that \cite{mohammed2017manifold} focuses on fitting a manifold in the noiseless case, which is out of the scope of this work.
The second one is counting on the regression idea or its variation to denoise points since locally the manifold can be parametrized by a function defined on the tangent space, including \cite{su2019OSonEEG,alagapan2019diffusion,NEURIPS2019_a76c0abe,sober2020manifold,aizenbud2021nonparametric,dunson2022inferring}.
While these algorithms vary from one to another in details, the common first step among all algorithms is {\em determining neighbors}, except the KDE based approach. Since the KDE approach is not feasible when the ambient space dimension is high, approaches in the line of KDE will not be discussed in the following paper, and we skip citing relevant papers.

Before proceeding, we shall mention that under the smoothness and compactness assumption of a manifold, among other quantities, we may classify the manifold denoising problem based on the following quantities---the rank of the support of the embedded manifold; that is, the rank of $\XX$, denoted by $D$, the ambient space dimension $p$, and the relationship between $p$, $d$ and the sample size $n$.
The rank assumption is determined by the isometric embedding $\iota$. Depending on $\iota$, we may need different strategies to denoise the manifold. The $n\epsilon^d$ condition describes how many points we have within a local ball. This condition also determines how to denoise the manifold. 

Our main contribution in this manuscript is providing a novel manifold denoiser combining optimal shrinkage (OS) \cite{nadakuditi2014optshrink} and ROSELAND \cite{shen2022robust,shen2022scalability}. For convenience, we call the algorithm {\em RObust and Scalable manifold Denoise with Optimal Shrinkage} (\textsf{ROSDOS}). 
The basic idea of \textsf{ROSDOS} is essentially a metric design algorithm. It depends on the fact that locally a $d$-dim manifold can be well approximated by a $d$-dim affine subspace and the following observation. Suppose we are able to find all points $x_{i_j}\in \mathcal{X}$, where $j=1,\ldots, n_i$, so that the associated clean companions $s_{i_j}$ satisfies $s_{i_j}\in B_\epsilon(s_i)$, where $B_\epsilon(x_i)$ is an Euclidean ball of a small radius $\epsilon>0$ centered at $s_i$, then we can recover $s_i$ by taking the mean or median of $x_{i_1},\ldots, x_{i_{n_i}}$ entrywisely. However, in practice it is challenging to locate $x_{i_1},\ldots, x_{i_{n_i}}$ from the noisy point cloud $\mathcal{X}$. Note that the usual $L^2$ norm is limited for this purpose, particularly when noise is large. Thus, we need a metric that allows us to determine $x_{i_1},\ldots, x_{i_{n_i}}$ from the noisy point cloud $\mathcal{X}$.

The manuscript is organized in the following way. In Section \ref{dm_setup}, we detail the mathematical model for the manifold denoising problem. In Section \ref{section:algorithm}, the proposed algorithm is given. 
In Section \ref{section:numerics}, a series of numerical results and comparisons with existing manifold denoisers are reported using synthetic datasets and a semi-real LFP dataset with high frequency electrical stimulus artifacts during deep brain stimulation. The theoretical support and more comparisons will be provided in the official paper of this manuscript.

Before ending this section, we shall mention that the current work could be viewed as a generalization of previous manifold denoising algorithm that has been applied to study the two-dimensional random tomography problem \cite{singer2013two}, recover the precision matrix \cite{gavish2019optimal}, extract the fetal electrocardiogram (ECG) from the trans-abdominal maternal ECG \cite{su2019OSonEEG}, remove the stimulation artifact from the intracranial electroencephalogram (EEG) \cite{alagapan2019diffusion}, denoise image manifolds \cite{pmlr-v9-gong10a}, and recently remove the high frequency electrical stimulus artifact in local field potential (LFP) recorded from subthalamic nucleus \cite{LFPOS2023}.

\section{Noisy manifold model in the high dimensional setup}\label{dm_setup}
  
Consider a $d$-dimensional compact smooth Riemannian manifold $(M, g)$ without boundary that is isometrically embedded in $\mathbb{R}^r$ via $\iota_r:M\hookrightarrow \mathbb{R}^r$. Denote $K=\max_{x,y}\|\iota_r(x)-\iota_r(y)\|_{\mathbb{R}^r}$ as the diameter of $M$.
Take a random vector $S:(\Omega,\mathcal{F},\mathbb{P})\rightarrow\mathbb{R}^{r}$ so that the range $S$ is supported on $\iota_r(M)$. Recall that according to Nash's isometric embedding theorem \cite{nash1956imbedding}, it is possible that $r\leq m(3m+11)/2$, but it does not exclude the possibility that $r>m(3m+11)/2$. 
$S$ induces a probability measure on $\iota_r(M)$, denoted by $\widetilde{\mathbb{P}}$. 
Without loss of generality, we assume that $\mathbb{E}S=0$; that is, the embedded manifold is centered at $0$.

\begin{Assumption}\label{Assumption:prob measure}
Fix a $d$-dimensional compact smooth Riemannian manifold $(M, g)$ without boundary that is isometrically embedded in $\mathbb{R}^r$ via $\iota_r:M\hookrightarrow \mathbb{R}^r$. 
Take a random vector $S:(\Omega,\mathcal{F},\mathbb{P})\rightarrow\mathbb{R}^{r}$ so that the range $S$ is supported on $\iota_r(M)$ and $\mathbb{E}S=0$.
Denote the induced probability measure on $\iota_r(M)$ induced by $S$ as $\widetilde{\mathbb{P}}$, which is assumed to be absolutely continuous with respect to the Riemannian measure on $\iota(M)$, denoted by $\iota_{\ast}dV(x)$. 
\end{Assumption}

Recall that according to Nash's isometric embedding theorem \cite{nash1956imbedding}, it is possible that $r\leq m(3m+11)/2$, but it does not exclude the possibility that $r>m(3m+11)/2$. Denote $K=\max_{x,y}\|\iota_r(x)-\iota_r(y)\|_{\mathbb{R}^r}$ as the diameter of $M$. 

\begin{Assumption}
	We call the non-negative function$\mathsf p$ defined on $M^{d}$ coming from $d\widetilde{\mathbb{P}}(x)=\mathsf p(\iota^{-1}(x))\iota_{\ast}dV(x)$ by the Radon-Nikodym theorem the probability density function (p.d.f.) associated with $X$. When $\mathsf p$ is constant, $X$ is called uniform; otherwise non-uniform. Assume $\mathsf p$ satisfies $\mathsf p \in \mathcal{C}^4(M)$ and $\inf_{x\in M^{d}}\mathsf p(x)>0$. 
\end{Assumption}

We consider the case when the ambient space dimension grows asymptotically as the dataset size $n$. To capture the manifold structure for the high dimensional dataset, we consider the following model.

\begin{Assumption}[High dimensional model]\label{Assumption manifoldH}
Take $p=p(n)$ so that $p\to \infty$ when $n\to \infty$. For each $p$, take an isometrical embedding $\bar{\iota}_p:\mathbb{R}^r\to \mathbb{R}^p$. Further, assume that $s_i$ are i.i.d. sampled from $\iota_p(M)$, where $\iota_p:=\bar{\iota}_p\circ \iota_r$. 
\end{Assumption}

Denote $\SX=[s_1,\ldots,s_n]\in \mathbb{R}^{p\times n}$ as the clean data matrix. Since the covariance matrix of $\SX$ is of full rank, the singular values of $\SX$ are nondegenerate almost surely when $n\to \infty$.
Note that there are several possibilities to model $\bar{\iota}_p$. For example, we can embed $\mathbb{R}^r$ into the first $r$ axes of $\mathbb{R}^p$ via $\bar{\iota}_p$, and sample the data. If the point cloud represents images, and the ambient space dimension represents the image resolution, then obviously the manifold representing the clean image is not embedded in only the first few axes.

\subsection{Noise model as a random matrix}
We now consider our noisy manifold model on top of Assumption \ref{Assumption manifoldH}.

\begin{Assumption}[Complex noise model]\label{Assumption Noise}
Let $A\in \mathbb{R}^{p\times p}$ and $B\in \mathbb{R}^{n\times n}$ be deterministic positive definite matrices with eigenvalues $\lambda_1^A \geq \lambda_2^A \geq \ldots \geq \lambda_p^A>0$ and  $\lambda_1^B \geq \lambda_2^B \geq \ldots \geq \lambda_n^B>0$ respectively and satisfy
\begin{equation}\label{ass3_eq2}
\sigma_1^A\vee  \sigma_1^B \le \tau^{-1}\ \mbox{ and } \ \pi_A([0,\tau])\vee  \pi_B([0,\tau]) \le 1 - \tau \,,
\end{equation}
where $\tau>0$ is a small constant, $\pi_A:=\frac{1}{p}\sum_{l=1}^p\delta_{\sigma_l^A}$ is the empirical spectral density of $A$ (similarly for $\pi_B$), and $\delta$ is the Dirac delta measure. Assume $\pi_A$ and $\pi_B$ converge as towards two compactly supported probability measures $\nu_1$ and $\nu_2$ as $n\to\infty$, where $\nu_1\neq \delta_0$ and $\nu_2\neq \delta_0$ and $\texttt{supp}(\nu_1)\cap(\mathbb{R}\backslash \{0\})$ and $\texttt{supp}(\nu_2)\cap(\mathbb{R}\backslash \{0\})$ contain $K_1$ and $K_2$ connected components.
The noisy dataset in the matrix form, $\XX\in \mathbb{R}^{p\times n}$, satisfies
\begin{align}\label{X=S+p^-a/2Xi}
\XX = \SX+p^{-\alpha}\boldsymbol\Xi\,,
\end{align}
where  $\alpha\geq0$, $\boldsymbol\Xi=A^{1/2}ZB^{1/2}=[\xi_1,\ldots,\xi_n]$ is independent of $S$ and $Z$ has independent entries with $\max_{i,j}|\mathbb{E}Z_{ij}|\leq n^{-2-\tau}$, $\max_{i,j}|\mathbb{E}[|Z_{ij}|^2]-n^{-1}|\leq n^{-2-\tau}$, and $\mathbb{E}[|\sqrt{n}Z_{ij}|^4]\leq C$ for some small constant $\tau>0$ and a constant $C>0$. 
\end{Assumption}

The assumptions of $A$, $B$ and $\boldsymbol\Xi$ are needed for the guaranteed performance of eOptShrink \cite{su2022cOS} when $p=p(n)$ and $p/n\to \beta\in [0,\infty)$, which is a critical part of the proposed \textsf{ROSDOS} algorithm. It is known that the spectral measure of $\boldsymbol\Xi\boldsymbol\Xi^\top$, denoted as $\mu_n$, converges in the distribution sense to a compactly supported probability measure $\mu$ \cite[Proposition 3.4]{couillet2014analysis} characterized by its Stieltjes transform and the number of connected components is bounded by $K_1K_2$ \cite[Proposition 3.3]{couillet2014analysis}. When $p/n\to 0$, it is shown in \cite[Proposition 2.1]{couillet2014analysis} that $\mu_n$ converges in the distribution sense to $\nu_1(m_2^{-1} dt)$, where $m_2:=\int s\nu_2(ds)$, as $n\to \infty$. See Figure \ref{fig:noiseSCspectrum} in Section \ref{section:numerics} for some examples. 

\begin{definition}
The dataset $\boldsymbol{X}$ satisfies the {\em low dimensional manifold with high dimensional complex noise} (LDM+HDCN) model if Assumptions \ref{Assumption:prob measure}-\ref{Assumption Noise} hold. When $p>1/2$, the noise is {\em negligible}, when $p=1/2$, the noise is {\em moderate}, and when $p<1/2$, the noise is {\em ample}.  
\end{definition}

By a direct calculation, $\texttt{cov}(\xi_i)=B_{ii}A/p^{2\alpha}$. Thus, $\texttt{tr}(\texttt{cov}(\xi_i))\leq \sigma_1^B\sum_{l=1}^p\sigma_1^A/p^{2\alpha}$. Since the spectra of $A$ and $B$  are both asymptotically compactly supported, the total energy of noise $\xi_i$ is of order $p^{1-2\alpha}$. The noise size is classified into {\em negligible}, {\em moderate} and {\em ample}, depending on if the total noise energy decreases to 0, converges to a constant, or blows up.

\begin{Assumption}[Local behavior]\label{Assumption LocalN}
In addition to $p$ in Assumption \ref{Assumption manifoldH}, take $\epsilon=\epsilon(n)$ so that $\epsilon\to 0$ and $n\epsilon^d\to \infty$, and $\frac{p}{n\epsilon^d}\to \beta\in [0,\infty)$ when $n\to \infty$. 
\end{Assumption}

Assumption \ref{Assumption LocalN} essentially says that $n\gg p$ when $\beta>0$, which may not always happen in practice. When $\beta=0$, we look at the relationship between $n$ and $p$. It is possible that $p/n\to \beta'\in (0,\infty)$ or $p/n\to \infty$. We need different strategies to handle these different cases.

\section{Proposed manifold denoise algorithm}\label{section:algorithm}

The key step of our proposed manifold denoiser, \textsf{ROSDOS}, is designing a high quality similarity metric $d$ so that 
\[
d(x_i,x_j)\approx d_g(s_i,s_j)\,,
\]
where $d_g$ is the geodesic distance on the manifold, particularly when $d_g(s_i,s_j)$ is small. This similarity metric allows us to find true neighbors, and hence an accurate recovery of the sample on the manifold.
The \textsf{ROSDOS} algorithm depends on the ROSELAND algorithm \cite{shen2022robust,shen2022scalability}, which is a computationally efficient variation of Diffusion Maps \cite{Coifman2006} and the extended optimal shrinkage (eOptShrink) algorithm \cite{su2022cOS}, which is an {\em optimal shrinkage} (OS) technique \cite{nadakuditi2014optshrink}. Both steps aim to handle noise by using the available geometric structure at hand. Below, we recall DM, ROSELAND and eOptShrink algorithms first, and then detail \textsf{ROSDOS} step by step with some practical considerations.

\subsection{Diffusion maps and ROSELAND in a nut shell} \label{sec:DM intro}

Suppose the dataset is $\mathcal{X}:=\{x_i\}_{i=1}^n\subset \mathbb{R}^p$. Fix a smooth kernel function $K$ that decays sufficiently fast and a bandwidth parameter $h>0$ chosen by the user. To simplify the discussion, we focus on the Gaussian kernel. Then, compute the affinity matrix $W^{(0)}\in \mathbb{R}^{n\times n}$ by
\begin{equation}\label{Definition affinity matrix W}
W^{(0)}_{ij} := 
\exp\left(-\frac{\|x_{i}-x_{j}\|^2_{\mathbb{R}^p}}{h}\right)(1-\delta_{ij})\,,
\end{equation}
where $\delta_{ij}$ is the Kronecker delta function,
and the corresponding {\em degree matrix} $D\in \mathbb{R}^{n\times n}$, which is a diagonal matrix defined as 
$D^{(0)}_{ii}:=\sum_{j=1}^nW^{(0)}_{ij}$. The idea of removing the diagonal entries and constructing the complete graph follows the suggestion in \cite[Theorem 3.1]{el2016graph}, which forces the random walk on the point cloud to be non-lazy and is particularly helpful when the noise level is high.
Then, define the {\em normalized} affinity matrix $W\in \mathbb{R}^{n\times n}$ \cite{Coifman2006} as 
\begin{equation}\label{Definition affinity matrix W}
W_{ij}:=\frac{W^{(0)}_{ij}}{D_{ii}^{(0)}D_{jj}^{(0)}}\,,
\end{equation}
where $W_{ij}$ is called the {\em normalized affinity} between $x_i$ and $x_j$.
With the normalized affinity matrix $W$, define the associated degree matrix $D\in \mathbb{R}^{n\times n}$ by
$D_{ii}:= \sum_{j=1}^nW_{ij}$.
The associated transition matrix associated with a random walk on the dataset $\mathcal{X}$ is defined as
\begin{equation} \label{Eq A ordinary DM}
A :=D^{-1}W\,.
\end{equation}
Since $A$ is similar to $D^{-1/2}W D^{-1/2}$ we can find its eigendecomposition with eigenvalues $1=\lambda_1>\lambda_2\geq\ldots\geq\lambda_n$ and the associated right eigenvectors $u_1,\ldots,u_n$. 
With the spectral decomposition of $A$, embedding dimension $q'$ and {\em diffusion time $t>0$}, the DM embeds $\mathcal{X}$ via the map
\begin{equation}\label{equation DM}
\Phi_t:x_{i}\mapsto e_i^\top {U}_{q'}L_{q'}^t 
\in \mathbb{R}^{q'},
\end{equation}
where ${U}_{q'}\in \mathbb{R}^{n\times q'}$ to be a matrix consisting of $u_2,\ldots,u_{q'+1}$, and $L_{q'}:=\texttt{diag}(\lambda_2,\ldots,\lambda_{q'+1})$. Then, define the diffusion distance (DD) by
\begin{equation}\label{definition DD metric}
d_{\texttt{DD}}({x}_i,{x}_j):=\|\Phi_t(x_{i})-\Phi_t(x_{j})\|\,.
\end{equation}
Recall that theoretically, under the manifold model, we have $d_{\texttt{DD}}(s_i,s_j)\approx d_g(s_i,\,s_j)$ when $s_i$ and $s_j$ are close. When $s_i$ and $s_j$ are far away, $d_{\texttt{DD}}(s_i,s_j)$ is large due to the spectral embedding property of DM. The robustness of DM and DD under independent noise has been well explored. See \cite{ElKaroui:2010a,el2016graph,9927456} for examples.

In practice, when $n$ is large, running DM is not efficient. We could consider the recently developed ROSELAND algorithm \cite{shen2022robust,shen2022scalability} to speed up this step. ROSELAND is a variation of DM in that $W$ is constructed in a different way. Select a subset of points $\mathcal{Y}=\{y_j\}_{j=1}^m\subset \mathcal{X}$, where $m=n^\gamma$ for $\gamma\in (0,1)$. Construct a landmark-affinity matrix $\bar W$ as
\[
\bar W_{ij}:=\exp\left(-\frac{\|x_{i}-y_{j}\|^2_{\mathbb{R}^p}}{h}\right)\,.
\]
Then build up a landmark-transition matrix $\bar A$ as
\[
\bar A:=\bar D^{-1/2}\bar W, \mbox{ where }\bar D:=\texttt{diag}(\bar W \bar W^\top \boldsymbol{1})
\]
and $\boldsymbol{1}\in \mathbb{R}^{n}$ is a vector with all entries $1$. 
Note that we could view $\bar W \bar W^\top$ as a quantity similar to the affinity matrix \eqref{Definition affinity matrix W}, while it is not constructed from a single kernel. It has been shown in \cite{shen2022scalability} that $\bar W \bar W^\top$ comes from an equivalent kernel that depends on the sample. 
Then apply the singular value decomposition (SVD) to $\bar A=\bar U \bar S \bar V^\top$. Note that by the relationship between the SVD of $\bar A$ and the EVD of $\bar A\bar A^\top=\bar D^{-1/2}\bar W \bar W\bar D^{-1/2}$, this SVD is equivalent to the EVD of $D^{-1/2}WD^{-1/2}$ in the original DM. 
With the SVD of $\bar A$, embedding dimension $q'$ and {\em diffusion time $t>0$}, the DM embeds $\mathcal{X}$ via the map
\begin{equation}\label{equation RL}
\bar\Phi_t:x_{i}\mapsto e_i^\top \bar D^{-1/2}\bar{U}_{q'}\bar S_{q'}^{2t} 
\in \mathbb{R}^{q'},
\end{equation}
where $\bar{U}_{q'}\in \mathbb{R}^{n\times q'}$ to be a matrix consisting of the second to $q'+1$ th left singular vectors, and $\bar L_{q'}$ is a diagonal matrix with the associated singular values in the diagonal entries. Then, define the ROSELAND DD by
\begin{equation}\label{definition DD metric}
d_{\texttt{ROSELAND}}({x}_i,{x}_j):=\|\bar\Phi_t(x_{i})-\bar\Phi_t(x_{j})\|\,.
\end{equation}
It has been shown in \cite{shen2022scalability} that asymptotically ROSELAND and ROSELAND DD behave like DM and DD under the clean manifold setup, particularly how the landmark impacts the final result. Its robustness to independent noise has also been shown in \cite{shen2022scalability}.

While in practice it is usually suggested to utilize the k nearest neighbors to construct the graph, we follow the suggestion in \cite[Theorem 3.1]{el2016graph} to consider the complete graph, particularly when the noise level is high. Note that the traditional $L^2$ norm is compromised by noise, especially when the dimensionality $p$ is high due to the concentration property. In this case, no neighboring information is trustable. Recall that $\|{x}_i-{x}_j\|^2=\|{s}_i-{s}_j\|^2+\|{z}_i-{z}_j\|^2+2\langle {z}_i-{z}_j,\, {s}_i-{s}_j\rangle$. While we need $\|{s}_i-{s}_j\|^2$ to construct the graph, we can only access $\|{x}_i-{x}_j\|^2$ in practice. Recall that if ${z}_i\sim N(0, I_p)$,  $\|{x}_i-{x}_j\|^2$ follows the the non-central chi-square distribution with mean $p+\|{s}_i-{s}_j\|^2$ and standard deviation $\sqrt{2p+4\|{s}_i-{s}_j\|^2}$. 
As a result, since $\|{s}_i-{s}_j\|^2$ is bounded under the LDM+HDCN model, it is dominated by noise when $p$ is large and the noise is ample, which misleads the neighboring information if we use $\|{x}_i-{x}_j\|^2$. In general, the distribution of $\|{z}_i-{z}_j\|^2$ is more complicated, but the same idea holds. We should mention that in MLOP \cite{faigenbaum2023manifold}, this high dimensional noise impact is noticed and a random projection idea is proposed to obtain a more accurate metric information.

\subsection{Optimal shrinkage in a nutshell}

Suppose we have a matrix 
\[
\XX=\begin{bmatrix} x_1 &  x_2 & \ldots & x_{n} \end{bmatrix}\in \mathbb{R}^{p\times (n_i+1)}
\]
that satisfies the noisy model detailed in the previous section holds.
Without loss of generality, we also assume $p\leq n$. If $p>n_i$, all the following operations can be done similarly by considering the transpose of $\XX$. 
Denote the singular value decomposition (SVD) of $\XX$ as 
\begin{equation}
    \XX=\sum^{p}_{l=1}\tilde{\sigma}_l \tilde{\boldsymbol u}_l\tilde{\boldsymbol v}_l^\top,
\end{equation}
where $\tilde{\sigma}_l$ is the $l$th singular value, $\tilde{\sigma}_1\geq \tilde{\sigma}_2\geq \ldots \geq\tilde{\sigma}_p$, and $\tilde{\boldsymbol u}_i\in O(p)$ and $\tilde{\boldsymbol v}_i\in O(n_i)$ are the corresponding left and right singular vectors.
Denote the eigenvalues of $\XX \XX^\top$ as $\tilde{\lambda}_i=\tilde{\sigma}_i^2$. The eOptShrink is carried out by evaluating the following quantities, one after the other.  
First, evaluate
\begin{equation}\label{equation lambda+} 
\hat{\lambda}_+:=\tilde{\lambda}_{[n^{1/4}]+1}+\frac{1}{2^{2/3}-1}(\tilde{\lambda}_{[n^{1/4}]+1}-\tilde{\lambda}_{2[n^{1/4}]+1}),
\end{equation}
where $[n^{1/4}]$ is the closest integer to $n^{1/4}$, and estimate the {\em effective rank} by
\begin{equation}\label{equation rhat}
    \hat{r}=\left|\left\{\tilde{\lambda}_i|\tilde{\lambda}_i>\hat{\lambda}_++n^{-1/3}\right\}\right|.
\end{equation}
Second, evaluate  
\begin{equation}\label{equation lambdaj} 
\hat{\lambda}_j:=\tilde{\lambda}_{k+1}+\frac{1-(\frac{j-1}{k})^{2/3}}{2^{2/3}-1}(\tilde{\lambda}_{2k+1}-\tilde{\lambda}_k+1)\,,
\end{equation}
where $k<p$ is a given parameter, and we set $k=10$ in this study. 
Third, derive
\begin{equation}\label{equation mhat}
    \hat{m}_{1,i}:=\frac{1}{p}\left(\sum_{j=1}^k\frac{1}{\hat{\lambda}_j-\tilde{\lambda}_i}+\sum_{j=k+1}^p\frac{1}{\tilde{\lambda}_j-\tilde{\lambda}_i}\right),\; \hat{m}_{2,i}:=\frac{1-\beta}{\tilde{\lambda}_i}+\beta\hat{m}_{1,i},
\end{equation} 
\begin{equation}\label{equation mhat'}
    \hat{m}_{1,i}':=\frac{1}{p}\left(\sum_{j=1}^k\frac{1}{(\hat{\lambda}_j-\tilde{\lambda}_i)^2}+\sum_{j=k+1}^p\frac{1}{(\tilde{\lambda}_j-\tilde{\lambda}_i)^2}\right),\;\hat{m}_{2,i}':=\frac{1-\beta}{\tilde{\lambda}_i^2}+\beta\hat{m}_{1,i},
\end{equation}
where $\beta=p/n$. Fourth, for $i=1,\ldots,\hat{r}$, evaluate
\begin{equation}\label{equation varphii'}
\hat{d}_i=\hat{\varphi}_i\sqrt{\hat{a}_{1,i}\hat{a}_{2,i}},
\end{equation}
where 
\begin{equation}\label{equation di'}
    \hat{\varphi}_i=\sqrt{\frac{1}{\hat{T}_i}},\;\hat{a}_{1,i}=\frac{\hat{m}_{1,i}}{\hat{\varphi}_i^2\hat{T}_i^{'}},\;\hat{a}_{2,i}=\frac{\hat{m}_{2,i}}{\hat{\varphi}_i^2\hat{T}_i^{'}},
\end{equation}
and 
\begin{equation}\label{equation Ti}
\hat{T}_i=\tilde{\lambda}_i\hat{m}_{1,i}\hat{m}_{2,i},\;\hat{T}_i^{'}=\hat{m}_{1,i}\hat{m}_{2,i}+\tilde{\lambda}_i\hat{m}_{1,i}'\hat{m}_{2,i}+\tilde{\lambda}_i\hat{m}_{1,i}\hat{m}_{2,i}'
\end{equation}
and calculate the optimal shrinked data matrix as 
\begin{equation}\label{definition Shat}
\hat{\SX}=\sum_{l=1}^{\hat{r}}\hat{d}_l\tilde{\boldsymbol u}_l\tilde{\boldsymbol v}_l^\top=\begin{bmatrix} \hat{{s}}_1 & \hat{{s}}_2 & \ldots & \hat{{s}}_{n_i} \end{bmatrix}.
\end{equation}
To speed up the algorithm, we can apply the randomized idea \cite{halko2011finding}.

The bulk edge is estimated in \eqref{equation lambda+}, which estimates the effective rank \eqref{equation rhat}. In practice, there is no access to the covariance structure of $\Xi$ and hence no access to its asymptotic spectral distribution, particularly the bulk. These equations give the desired information. 
The optimal singular value deformation for the denosing purpose is achieved by
\eqref{equation lambdaj}, \eqref{equation mhat}, \eqref{equation mhat'}, \eqref{equation di'} and \eqref{equation Ti}, which establish the relationship between the clean singular values and singular vectors and noisy singular values and singular vectors. This relationship leads to the optimal shrinkage of the singular values in \eqref{equation varphii'}, and hence the optimal recovery of the stimulus artifact matrix \eqref{definition Shat}.

Recall that when the rank is small in a noisy matrix $\XX\in \mathbb{R}^{p\times n}$, where $p=p(n)$ so that $p/n\rightarrow \beta$ when $n\to \infty$ with $0<\beta\leq1$ (when $\beta>1$, the matrix can be simply transposed so that $\beta\leq 1$ is fulfilled), the singular values and singular vectors of $\XX$ are both biased estimators \cite{benaych2012singular}. In fact, if a singular value is larger than a threshold $\alpha_+$ determined by $\beta$ and the spectral structure of $\Xi$, the singular value and the associated singular vector are both biased; if a singular value is smaller than the threshold, it is buried by the noise and cannot be recovered. The number of sufficiently large singular values is called the {\em effective rank}.
This bias downgrades the performance of the commonly used TSVD, even if all singular values are above the threshold, and the downgrade gets worse as $\beta$ increases.
One solution to this challenge is the OS technique, which {\em nonlinearly} corrects the singular values under some moment conditions of $\Xi$ so that the denoised matrix is optimal in some sense. If $\XX=\sum_{i=1}^{p}\tilde d_i\tilde{\boldsymbol u}_i\tilde{\boldsymbol v}_i^\top$, the OS of $\XX$ is $\check{\SX}=\sum_{i=1}^{p}\eta(\tilde d_i)\tilde{\boldsymbol u}_i\tilde{\boldsymbol v}_i^\top$, where $\eta:[0,\infty)\to [0,\infty)$ is a nonlinear function depending on how the ``optimal'' is measured. Usually, we ask for an $\eta$ that minimizes $\|\check{\SX}-\SX\|_F$, where $\|\cdot\|_F$ is the Frobenius norm. Other norms could be considered.
One realization of this OS idea is OptShrink  \cite{nadakuditi2014optshrink}, which assumes the knowledge of rank and that asymptotically the spectrum of $\Xi$ is compactly support with further assumptions of the spectral bulk edges and delocalization of singular vectors. The algorithms is simplified in \cite{gavish2017} when $\Xi$ contains i.i.d. entries with mild fourth moment condition. ScreeNOT \cite{donoho2020screenot} and the whitening idea \cite{gavish2022matrix} are other OS algorithms, but inaccessible knowledge, like the rank information or part of the covariance structure of $\Xi$, is needed. Clearly, $\Xi$ considered in \eqref{definition ZX} does not fulfill the i.i.d. entries assumption, the covariance structure of $\Xi$ and the rank of $\SX$ is usually unknown, so methods from \cite{nadakuditi2014optshrink,gavish2017,donoho2020screenot,gavish2022matrix} are not applicable in our problem.
The eOptShrink algorithm \cite{su2022cOS} is a solution to these limitations. It is an extension of OptShrink when $\Xi$ satisfies the separable covariance condition. We refer readers with interest in technical details to \cite{su2022cOS}.

\subsection{\textsf{ROSDOS} algorithm}

Now we are ready to detail the \textsf{ROSDOS} algorithm step by step.

\subsubsection{Step 1: Global similarity metric design} \label{sec:OKTS_1stOS}

In the first case that the subspace hosting the embedded manifold is high and the condition number of the embedding is high, we apply DM and embed $\mathcal{X}$ via the map $\Phi_t$. The global similarity metric is defined as
\begin{equation}
d_{\texttt{global}}(x_i,\,x_j):=d_{\texttt{ROSELAND}}(x_i,\,x_j)
\end{equation}
When $p/n\to \gamma\in(0,\infty)$, we need an extra condition that the noise level depends on $p$ and is asymptotically small.

In the second case that the subspace hosting the embedded manifold is high and the condition number of the embedding is small, or if the subspace hosting the embedded manifold is low, we propose to directly apply eOptShrink to denoise $\mathcal{X}$; that is, denote the denoised matrix as $\hat{\SX}$ as in \eqref{definition Shat}. Define
\begin{equation}
d_{\texttt{global}}(x_i,\,x_j):=\|\hat{s}_i-\hat{s}_j\|_2\,.
\end{equation}
Recall that when $p/n\to 0$, eOptShrink is reduced to the traditional TSVD.
When $p/n\to \gamma\in(0,\infty)$, we only apply Step 1 with eOptShrink and jump to Step 3 without running Step 2.

\subsubsection{Step 2: local similarity metric design}
Fix $x_i$. 
Note in the ideal situation, all segments close to $x_i$ in the sense of $d_{\texttt{global}}$ should be real neighbors of $x_i$. However, in practice it is not always the case due to the noise. We thus carry out this second step to more accurately determine neighbors.

Find the $K$ nearest neighbors of $x_i$ measured by $d_{\texttt{global}}$, denoted as $ x_{i_1},\ldots, x_{i_K}$, and form a matrix 
\[
\XX_i=\begin{bmatrix} x_i &  x_{i_1} & \ldots & x_{i_K} \end{bmatrix}\in \mathbb{R}^{p\times (K+1)}. 
\]
We take $k$ small so that if $s_{i_1},\ldots,s_{i_K}$ are {\em true} neighbors of $s_i$, they are supported in a small ball centered at $s_i$ so that these $k$ points can be well approximated by a $d$-dim affine subspace. 
Then, apply eOptShrink to $\XX_i$ and denote the optimal shrinked data matrix as 
$\hat{\SX}_i$, and define the {\em local similarity metric} 
\begin{equation}\label{definition dOS metric}
d_{\texttt{local}}({x}_i,{x}_{i_j}):=\|\hat{s}_i-\hat{s}_{i_j}\|_2\,,
\end{equation}

\subsubsection{Step 3: Recover the manifold}

For each $i$, find $k\in \mathbb{N}$ nearest neighbors in ${\XX}_i$ by $d_{\texttt{OS}}$, where $k<K$ is chosen by the user. 
Finally, $s_i$ is recovered by taking the entrywise median of those $k$ chosen columns in ${\XX}_i$, and we denote the result as $\tilde{s}_i$. In principle, we could also consider mean, but median is suggested since it is less impacted by outliers, which is common in practice.

\subsection{Some technical considerations}

We elaborate some technical details of \textsf{ROSDOS}. If $r$ in \eqref{SX model clean} is small, eOptShrink in Step 1 essentially behaves like truncated SVD (TSVD) \cite{golub1965calculating,golub1987generalization}.
If $r$ is high, or even $r=p$ and the singular values decay to zero fast, eOptShrink would keep the strong singular values of $\SX$, technically those higher than the threshold $\alpha_+$. As a result, the denoised data matrix $\hat{\SX}$ is close to the signal matrix $\SX$ in the Frobenius norm sense, but details with small singular values are sacrificed. These details with small singular values are associated with high frequency content of the signal. This is the case in the LFP example. Using the manifold language, applying eOptShrink in Step 1 is projecting the embedded manifold into a lower dimensional subspace and ``expect'' that the embedded manifold after project is diffeomorphic to the true manifold. If the embedded manifold after project is diffeomorphic to the true manifold, we reduce the noise impact so that we can obtain better neighbor information, but the metric is deformed. Note that in general projection might destroy the topology of the manifold. 
In the worst case that $r=p$ and the singular values are all fixed (no decay), eOptShrink does not work.
Compared with eOptShrink, the benefit of applying DM in Step 1 is threefold. First, it is robust to noise (see below for a summary of existing results and new result). Second, it allows an almost isometric embedding of the underlying manifold. Third, in the worst case when $r=p$ and the condition number of embedding is large, as long as there is a low dimensional structure, DM and DD could help determine neighbors. 
The application of eOptShrink in Step 2 is tailored to handle the error remains in Step 1 by taking the nonlinear structure of the low dimensional manifold into account, if we have sufficient data points. 
Indeed, with the smooth manifold assumption, around $s_i$ locally the $d$-dim manifold could be well approximated by a $d$ dimensional affine subspace, and hence $\XX_i$ can be well modeled by a rank $r'$ data matrix, where $r'\leq r$ but only the first $d$ singular values are asymptotically dominant; that is, the effective rank is $d$. By Assumption \ref{Assumption LocalN}, the application of eOptShrink to $\XX_i$ in Step 2 leads to the optimal recovery of $s_i$. If we have much more points so that locally the available points is much larger than $p$, the eOptShrink is reduced to TSVD.

\section{Numerical result}\label{section:numerics}

We compare the proposed \textsf{ROSDOS} with the following manifold denoise or manifold fitting algorithms, FPM \cite{pmlr-v75-fefferman18a}\footnote{\url{https://github.com/zhigang-yao/manifold-fitting}}, NRPCA \cite{NEURIPS2019_a76c0abe}\footnote{\url{https://github.com/rrwng/NRPCA}}, MMLS \cite{aizenbud2021nonparametric}\footnote{\url{https://github.com/aizeny/manapprox}}, MrGap \cite{dunson2022inferring}\footnote{\url{https://github.com/wunan3/Manifold-reconstruction-via-Gaussian-processes}} and mFPM \cite{yao2023manifold}\footnote{\url{https://github.com/zhigang-yao/manifold-fitting}}. We skip the comparison with ridge extraction based on KDE approach since it is not suitable for high dimensional data. We also skip the comparison with MLOP and principle manifold approach since they are not suitable for the manifold denoising mission. How to modify these algorithms to denoise manifold is out of the scope of this paper. The Matlab code to reproduce results in this section can be found in \url{XX}.

\subsection{Numerical details of different algorithms}

For MMLS \cite{aizenbud2021nonparametric}, FPM \cite{pmlr-v75-fefferman18a}, NRPCA \cite{NEURIPS2019_a76c0abe}, MrGap \cite{dunson2022inferring} and mFPM \cite{yao2023manifold}, the common first step is estimating the local geometric structure, like tangent space or normal space, for the manifold denoise. We run these algorithms with the true dimension if the dataset is simulated, and with the estimated dimension from the clean dataset. 
We determine the local ball size using the clean dataset. Specifically, we find a radius $r$ so that $75\%$ of noisy points satisfy the following property. For any of these $75\%$ noisy points, the ball centered at this noisy point with the radius $r$ contains at least $k$ clean points, where $k\in \mathbb{N}$ is chosen from $\{25, 50,100,200\}$ so that the performance is the best. This is to guarantee that the ball centered at a noisy point overlaps with the clean manifold. 
For MrGap, the Gaussian process depends on three parameters, $A$, $\rho$ and $\sigma$, where $A$ and $\rho$ is for the Gaussian covariance structure and $\sigma$ is the noise strength. We search for the optimal $A$, $\rho$ and $\sigma$ over $(\exp(-4),\exp(6))$, $(\exp(-4),\exp(8))$ and $(\exp(-4),\exp(6))$, each of them is discretized into $18$ log-linear grid. 
For NRPCA, since we do not consider the sparse outliers, we only call the denoising part of the algorithm. The noise level is input to the algorithm and the optimal performance is searched over different numbers of nearest neighbors, including the suggested $20$, $50$ and $100$. The other parameters for iterations are kept the same from the available Matlab code. 
For MMLS \cite{aizenbud2021nonparametric}, the noise level $\sigma$ is selected by the algorithm and the other parameters are kept the same from the available Python code. 

We define the {\em manifold signal-to-noise ration} (mSNR) from the perspective of viewing the manifold as the signal to quantify the relationship between clean signal and noise. Denote the covariance of the signal matrix $\boldsymbol{S}$ as $C_s$, and the covariance of the noise matrix $\boldsymbol{\Xi}$ as $C_\xi$. The mSNR is defined as $10\log_{10}\frac{\texttt{tr}(C_s)}{\texttt{tr}(C_\xi)}$. Note that this mSNR reflects the signal to noise relationship in the global sense without capturing the local details, and it only provides a rough guidance about how bad the signal is contaminated by noise. 

To compare the performance of different algorithms, we consider the following indices. Since our main practical application is recovering each sample point on the manifold, we focus on the pointwise standard normalized root mean squared error (NRMSE) defined as $\|\tilde{s}_i-s_i\|_2/\|s_i\|_2$, for $i=1,\ldots,n$.
To compare the performance of different algorithms, the Wilcoxon signed rank test is applied. The p-value $0.005$ is viewed as statistically significance \cite{benjamin2018redefine}. Bonferroni correction is applied to handle the multiple test.

The entirety of computational tasks below is executed using Matlab R2021b on a MacBook Pro from the year 2017, operating on macOS Monterey 12.1, without the utilization of parallel computation capabilities. 
The code reproducing results this section will be announced in Github in the official paper of this manuscript.

\subsection{Simulated dataset}
The first manifold, $M_1$, we consider is the following 1-dim manifold embedded in $\mathbb{R}^p$ via
\begin{align*}
\iota_1: \theta\in  [0,2\pi)\mapsto [&\sin(\theta) \  \cos(\theta)/2 \  \sin(2\theta)/3 \   \cos(2\theta)/4 \ldots\\
&    \sin(J\theta)/(2J-1) \  \cos(J\theta)/(2J) \   0 \  \ldots \ 0  ]^\top \in \mathbb{R}^{p}
\end{align*}
with $J=\lceil 2p/5\rceil$. 
Note that since $\sin(k\theta)$ and $\sin(l\theta)$ are orthogonal, the $1$-dim manifold $M_1$ occupies the first $2\lceil 2p/5\rceil$ linear subspace of $\mathbb{R}^{p}$. Denote $X^{(1)}=\iota_1\circ X_1$, where $X_1\sim U(0,2\pi)$, to be a random vector of interest. 
The second manifold, $M_2$, is the 2-dim random tomography dataset \cite{singer2013two}, which is a diffeomorphic embedding of the canonical $S^1\subset \mathbb{R}^2$ to $\mathbb{R}^p$ via $\iota_2$. Denote $X^{(2)}=\iota_2\circ X_2$, where $X_2$ is a random vector with uniform distribution over the canonical $S^1$, to be the second random vector of interest.
The third manifold, $M_3$, is the Klein bottle embedded in the first four axes of $\mathbb{R}^p$ via 
\[
\iota_3: (t,s)\in [0,2\pi)\times [0,2\pi)\mapsto \begin{bmatrix}
(2\cos(t)+1)\cos(s)\\ (2\cos(t)+1)\sin(s)\\ 2\sin(t)\cos(s/2)\\ 2\sin(t)\sin(s/2)\\ 0 \\\vdots \\ 0
\end{bmatrix}\in \mathbb{R}^p\,.
\]
Denote $X^{(3)}=\iota_3\circ X_3$, where $X_3$ is a random vector with uniform distribution over $[0,2\pi)\times [0,2\pi)\subset \mathbb{R}^2$, to be the third random vector of interest.

\begin{figure}[!hbt]
    \centering
        \includegraphics[width=0.32\textwidth]{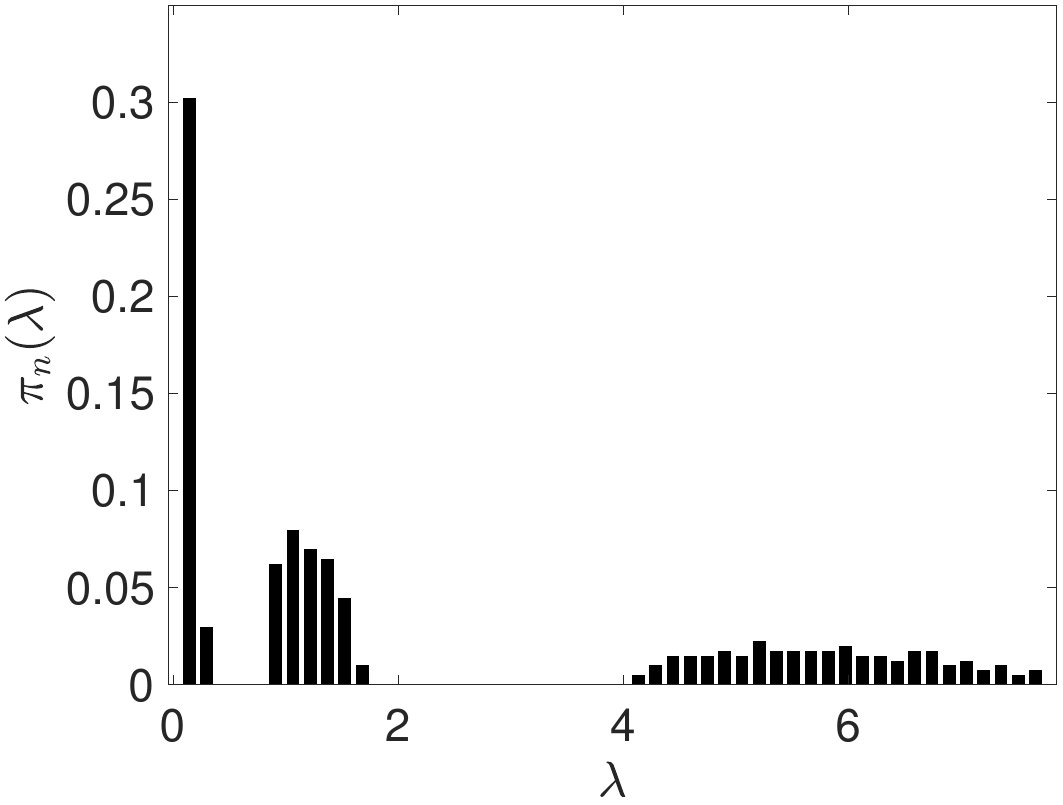} 
        \includegraphics[width=0.32\textwidth]{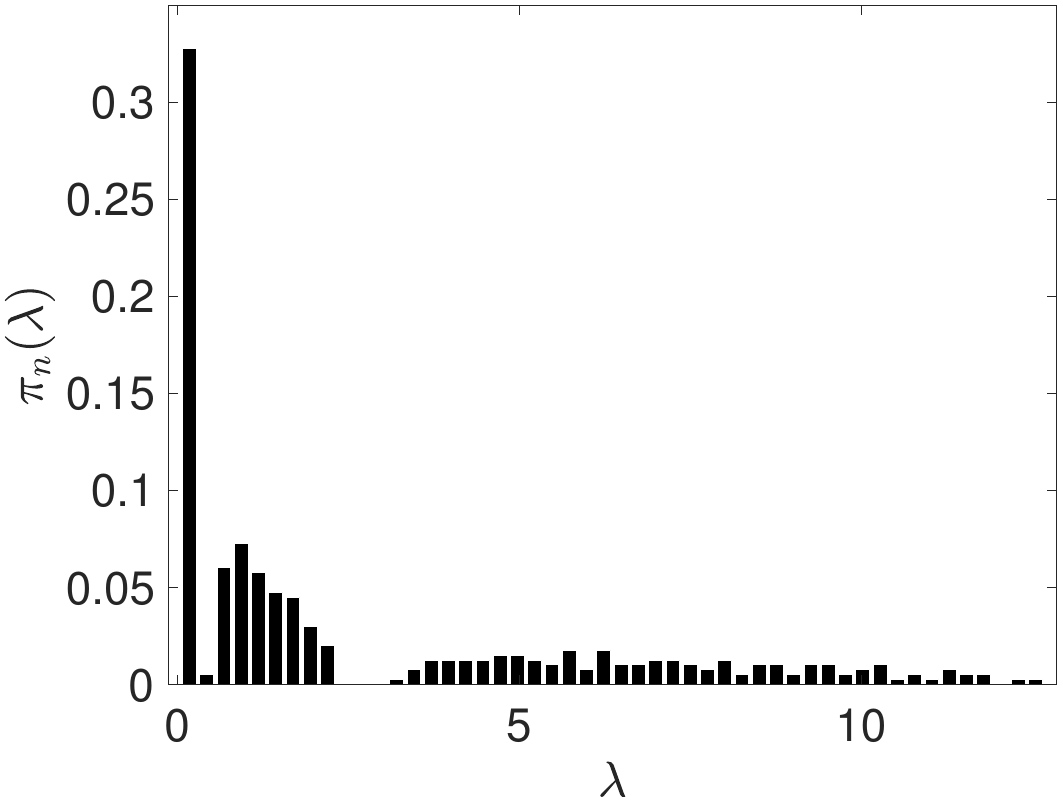} 
        \includegraphics[width=0.32\textwidth]{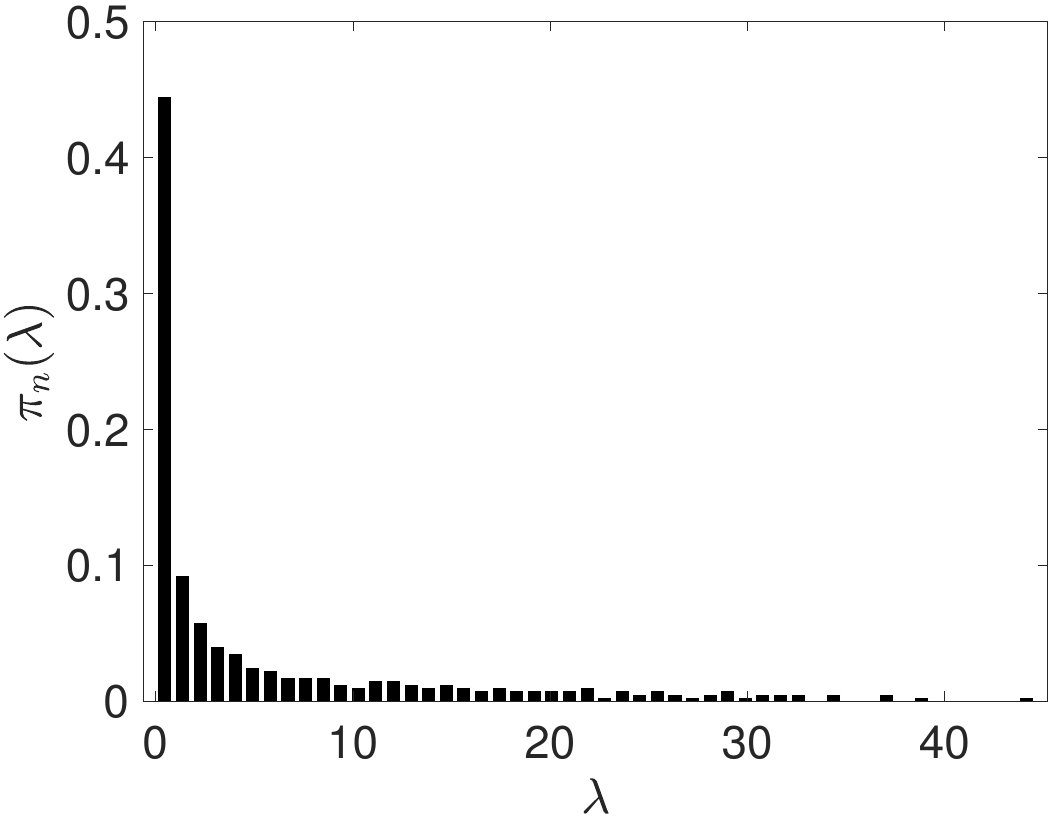} 
    \caption{\label{fig:noiseSCspectrum}The empirical spectral density of $\boldsymbol\Xi_1\boldsymbol\Xi_1^\top$. Left: $p=400$ and $n=10000$; middle: $p=400$ and $n=2000$; right: $p=400$ and $n=600$. }
\end{figure}

We generate two types of noise. The first one is the standard Gaussian noise, where $\boldsymbol{\Xi}_{0}=X$ has i.i.d. entries following $N(0, 1)$, where $\alpha>0$. The second noise has a separable covariance structure satisfying $\boldsymbol{\Xi}_{1}=A^{1/2}XB^{1/2}$, where $A$, $X$ and $B$ are generated in the following way. 
Construct $\lambda_j=1+l_j/32$ for $1\leq j\leq \lfloor p/3\rfloor$, $\lambda_j=1/4+l_j/32$ for $\lfloor p/3\rfloor+1\leq j\leq \lfloor 2p/3\rfloor$ and $\lambda_j=1/2+l_j/32$ for $\lfloor 2p/3\rfloor+1\leq j\leq p$, where $l_j$ is the eigenvalues of the symmetric random matrix with i.i.d. entries following $N(0,1/p)$; that is, $l_j$ follows the Wigner semi-circle law with the radius $1$. Then, randomly pick a $Q\in O(p)$ and set $A:=Q\texttt{diag}[\lambda_1,\ldots,\lambda_p]Q^\top$. 
For $B$, construct $\ell_1,\ldots,\ell_{\lfloor n/2\rfloor}$ by independently sampling from $U(0,1)/4+1/6$, and $\ell_{\lfloor n/2\rfloor+1},\ldots,\ell_n$ by independently sampling from $T_6/8+1$, where $T_6$ is the student T distribution of degree 6. Then, randomly pick a $\bar Q\in O(n)$ and set $A:=\bar Q\texttt{diag}[\ell_1,\ldots,\ell_p]\bar Q^\top$.
For $X\in \mathbb{R}^{p\times n}$, it has i.i.d. entries following $\tau$, where $\tau$ is the student T distribution of degree 5 with a normalized standard deviation $1$. See Figure \ref{fig:noiseSCspectrum} for an illustration of the spectral distribution of $\boldsymbol\Xi_1\boldsymbol\Xi_1^\top$.

With the above preparation, we consider the following noisy datasets. 
For $i=1,2,3$, sample $n$ points i.i.d. from $X^{(i)}$, and denote the resulting clean dataset $\mathcal{S}^{(i)}=\{ x^{(i)}_{l}\}_{l=1}^{n}\subset \mathbb{R}^{p}$ and the associated clear data matrix $\mathbf{S}^{(i)}\in \mathbb{R}^{p\times n}$; that is, the $l$-th column of $\mathbf{X}_i$ is $x_{i,l}$. 
Construct noisy data matrix $\boldsymbol{X}^{(i,j,\alpha)}=\mathbf{S}^{(i)}+\boldsymbol{\Xi}_j/p^{\alpha}$, where $\alpha=1,1/2,1/3$, $i=1,2,3$ and $j=0,1$; that is, we consider 18 different noisy datasets.
Below, set $p=200$ and $n=5,000$.

\begin{figure}[!hbt]
    \centering
        \includegraphics[width=0.32\textwidth]{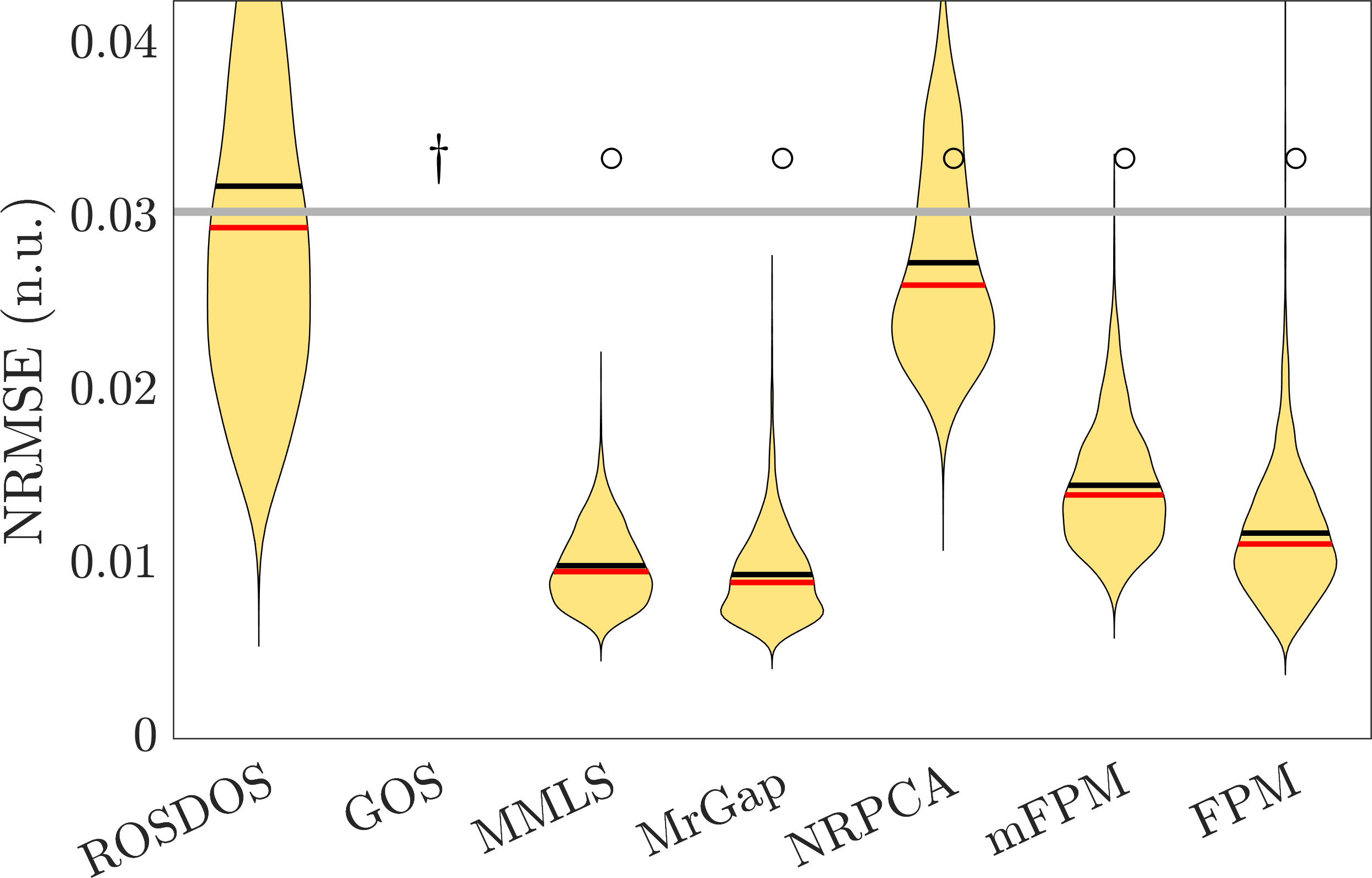} 
        \includegraphics[width=0.32\textwidth]{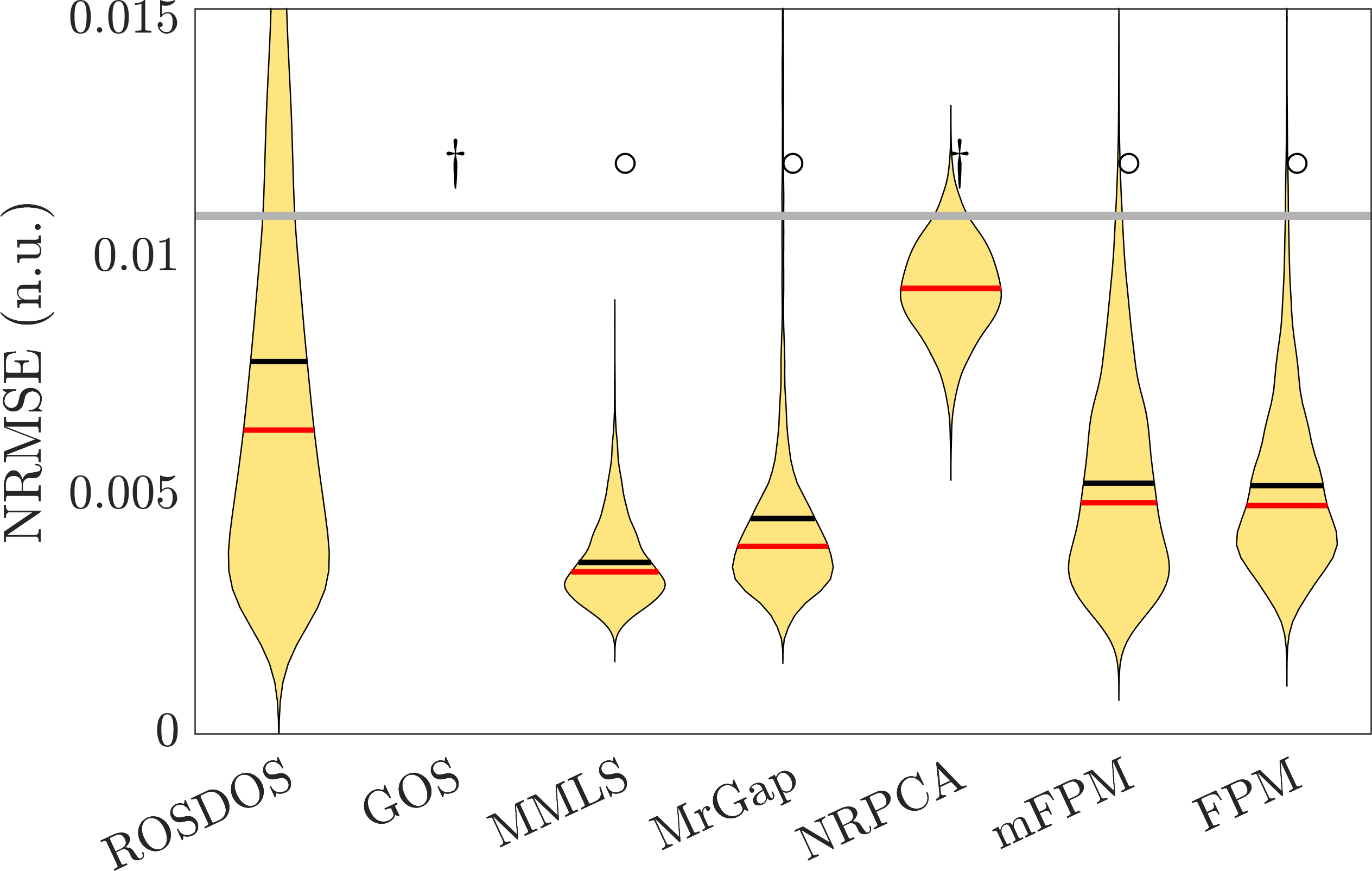} 
        \includegraphics[width=0.32\textwidth]{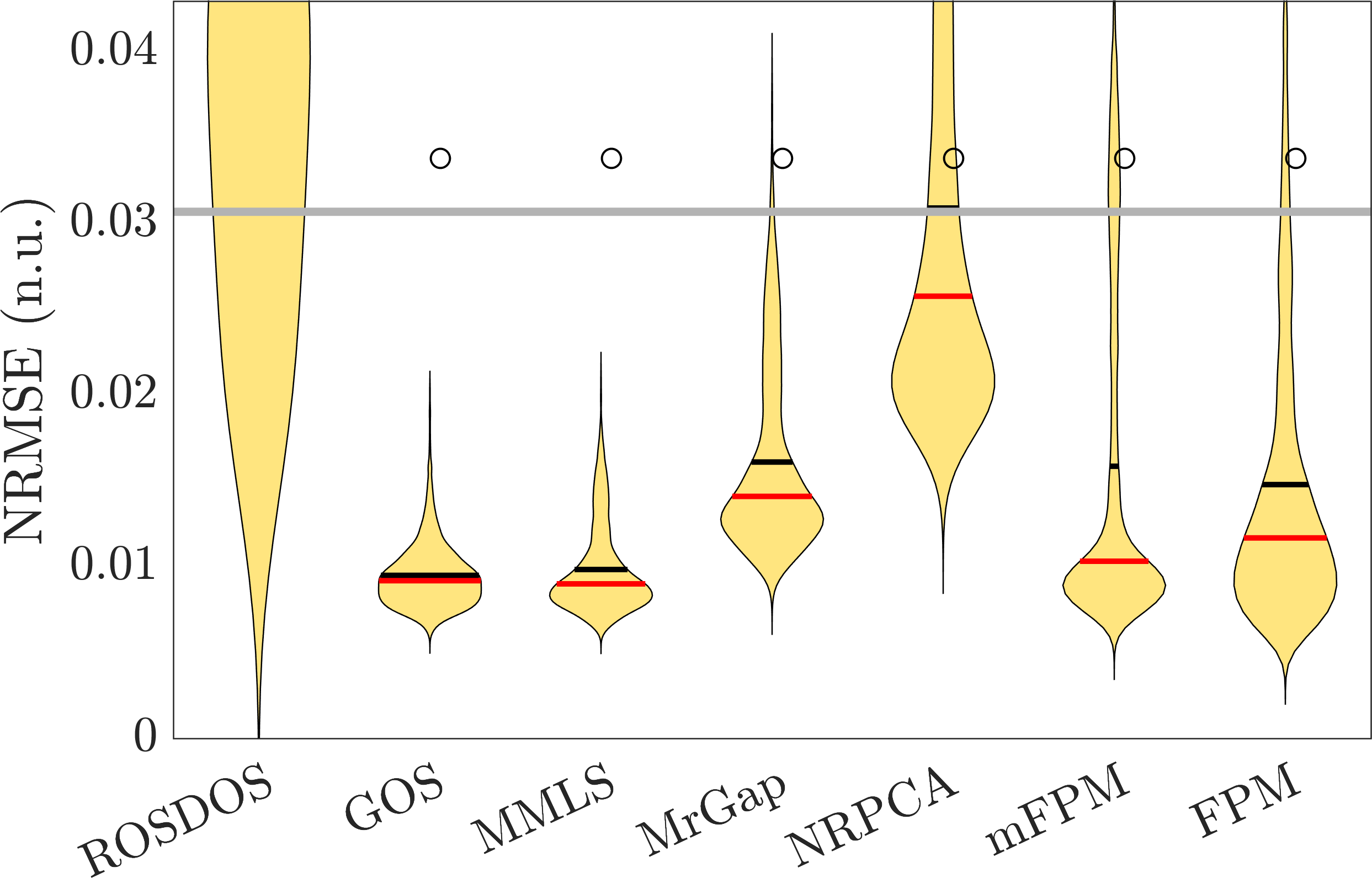} \\
        \includegraphics[width=0.32\textwidth]{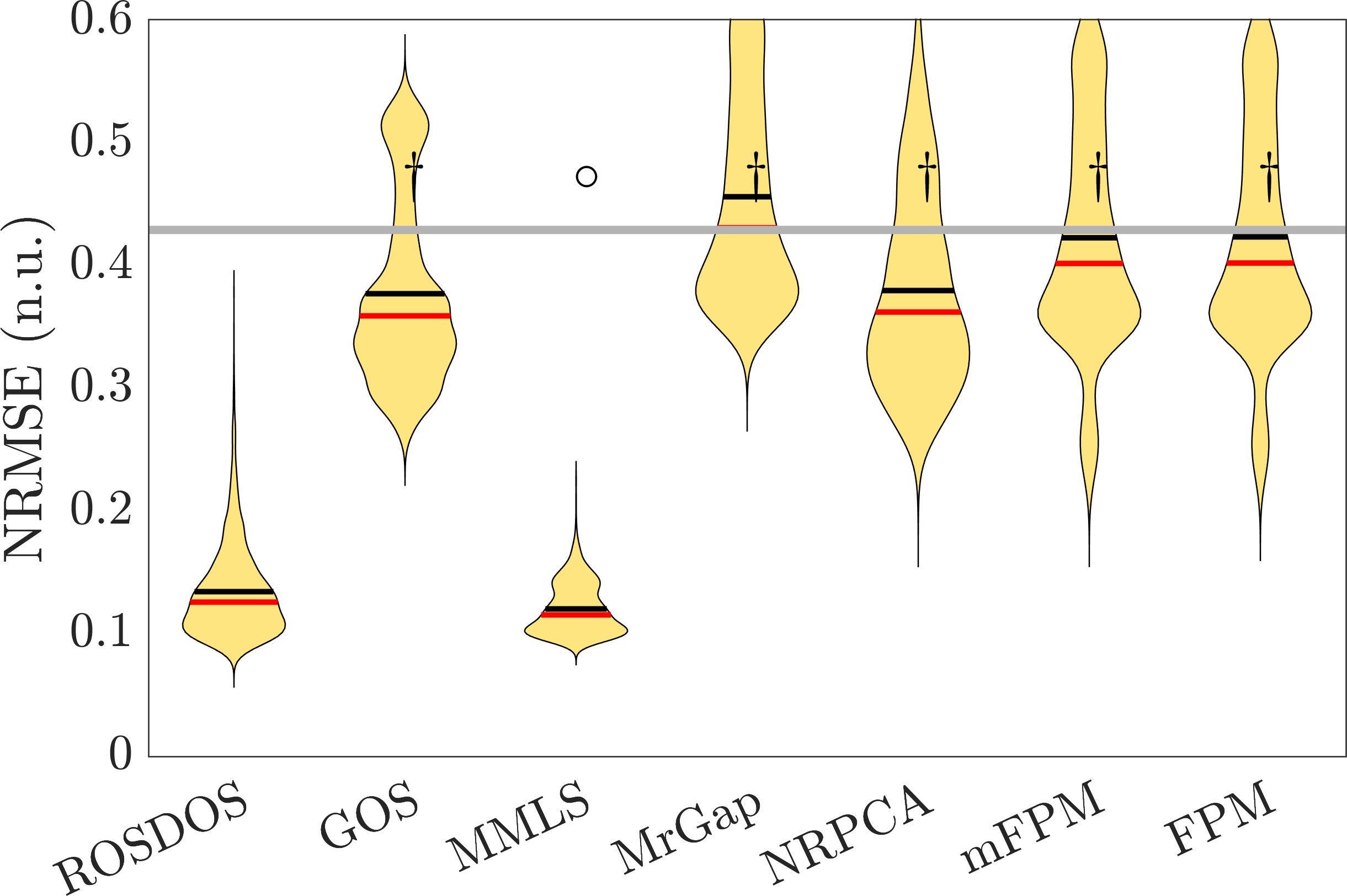} 
        \includegraphics[width=0.32\textwidth]{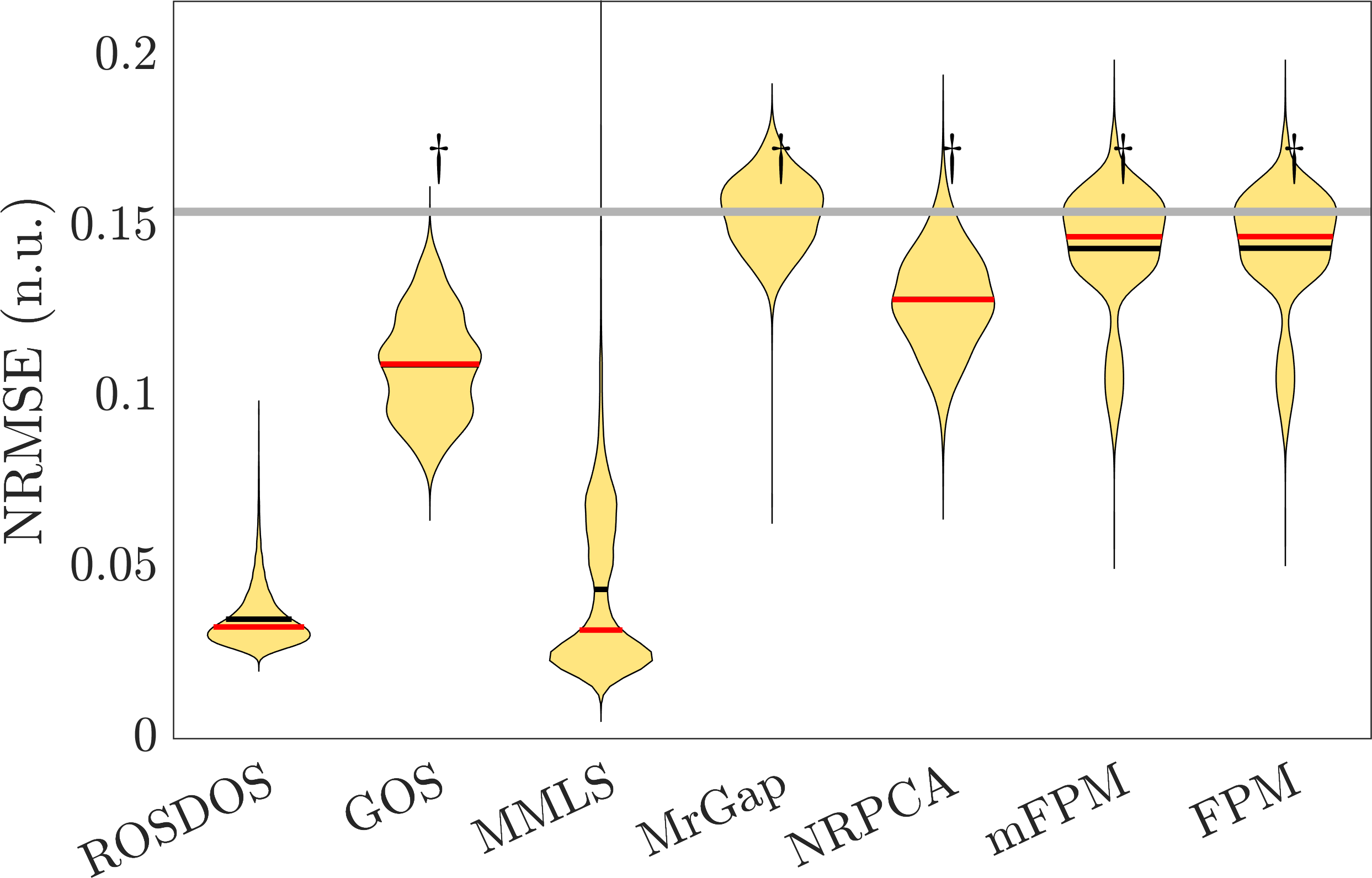} 
        \includegraphics[width=0.32\textwidth]{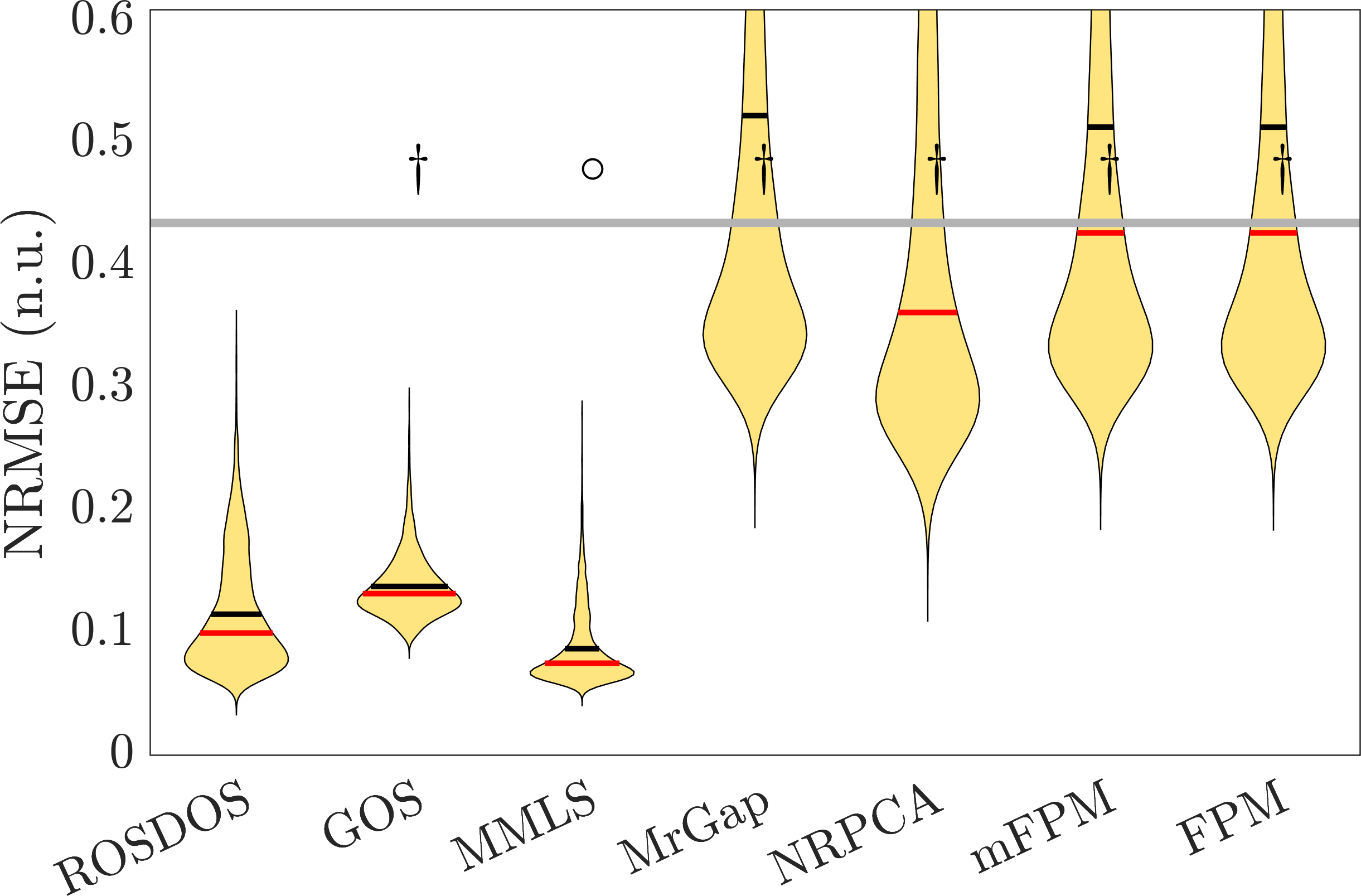} \\
        \includegraphics[width=0.32\textwidth]{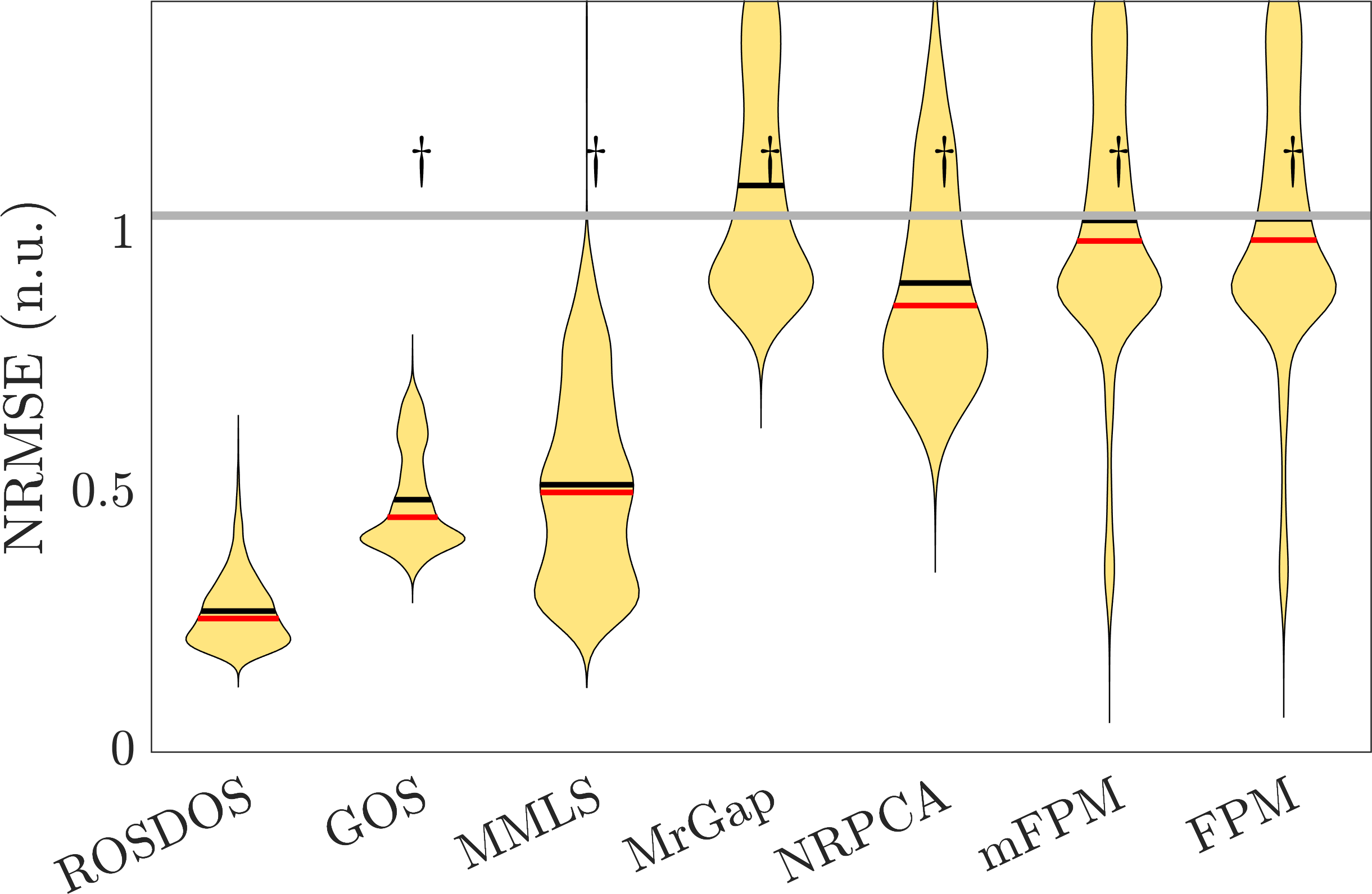} 
        \includegraphics[width=0.32\textwidth]{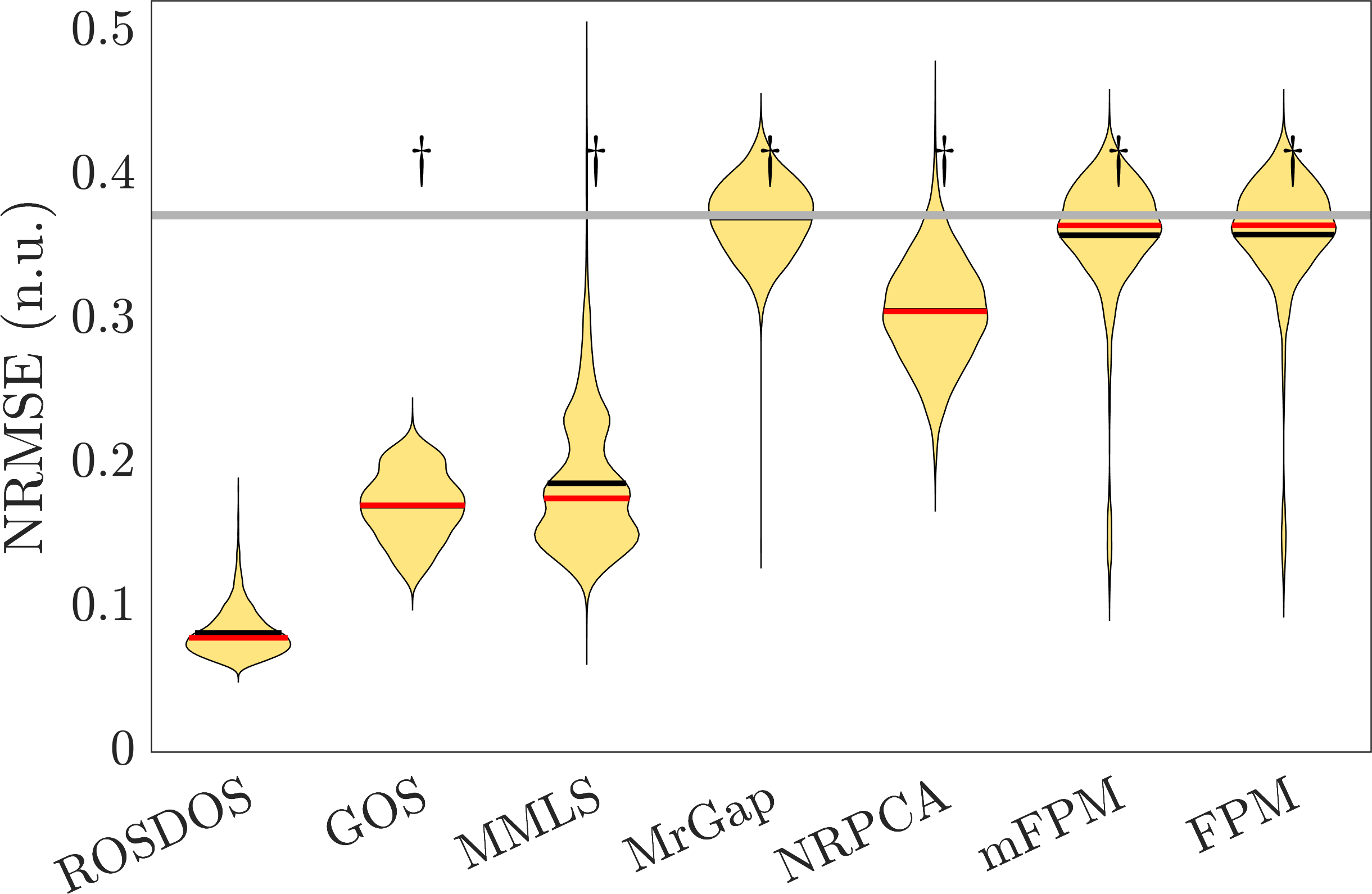} 
        \includegraphics[width=0.32\textwidth]{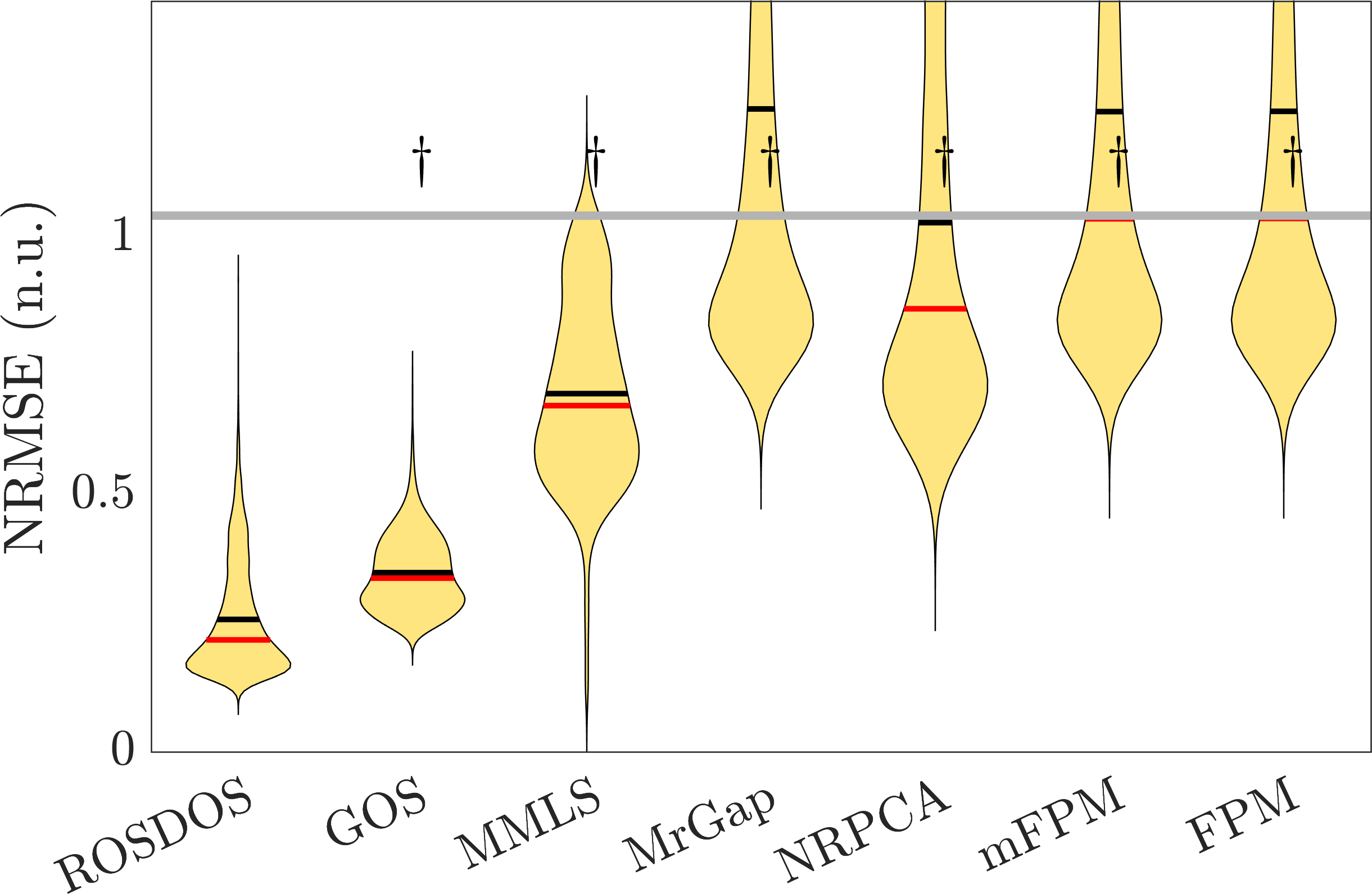} 
    \caption{\label{fig:SimuNRMSE}A summary of denoising efficiency of different algorithms over $12$ different simulated databases with Gaussian noise in terms of NRMSE. The distributions of NRMSE of ROSDOS, GOS, MMLS, MrGap, NRPCA, mFPM and FPM are shown in yellow. From left to right columns, the manifolds are $M_1$, $M_2$ and $M_3$. From the first to third rows are associated with $\alpha=1, 1/2$ and $1/3$. For each method, the distribution of NRMSE is estimated using the kernel density estimation with the Gaussian kernel with the optimal kernel bandwidth, which is shown as the violin plot. The gray horizontal line is the median of $\{\|\xi_i\|_2/\|s_i\|_2\}$. The red bar indicates the median and the black bar indicates the mean. To enhance the visualization, the y-axis upper bound is set to $1.4$ times of the median of $\{\|\xi_i\|_2/\|s_i\|_2\}$. The dagger (circle respectively) indicates that \textsf{ROSDOS} performs better (worse respectively) than the algorithm under comparison. The dagger (circle respectively) indicates that \textsf{ROSDOS} performs better (worse respectively) than other manifold denoiser.
    }
\end{figure}

\begin{figure}[!hbt]
    \centering
 	\includegraphics[width=0.32\textwidth]{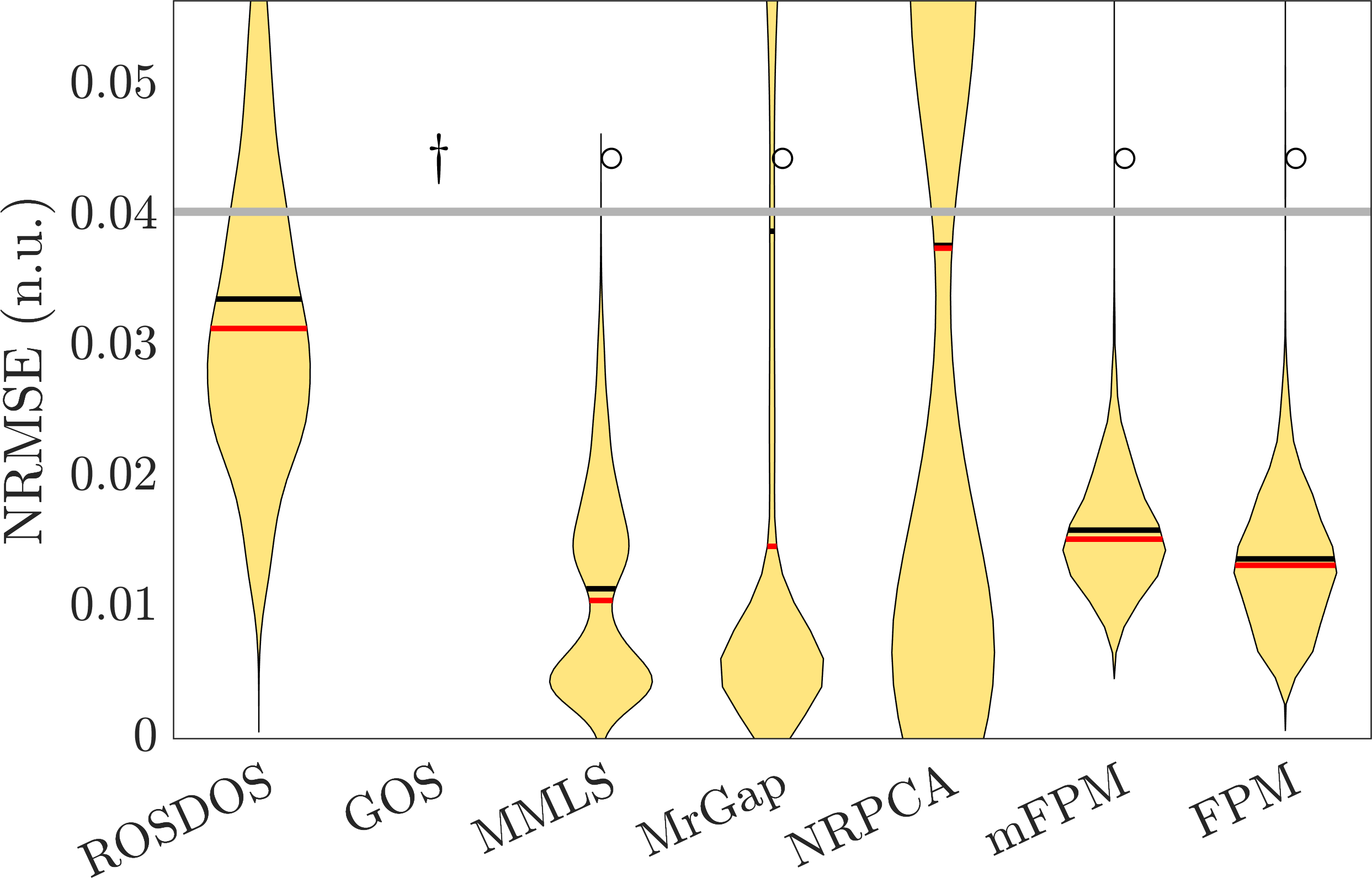} 
        \includegraphics[width=0.32\textwidth]{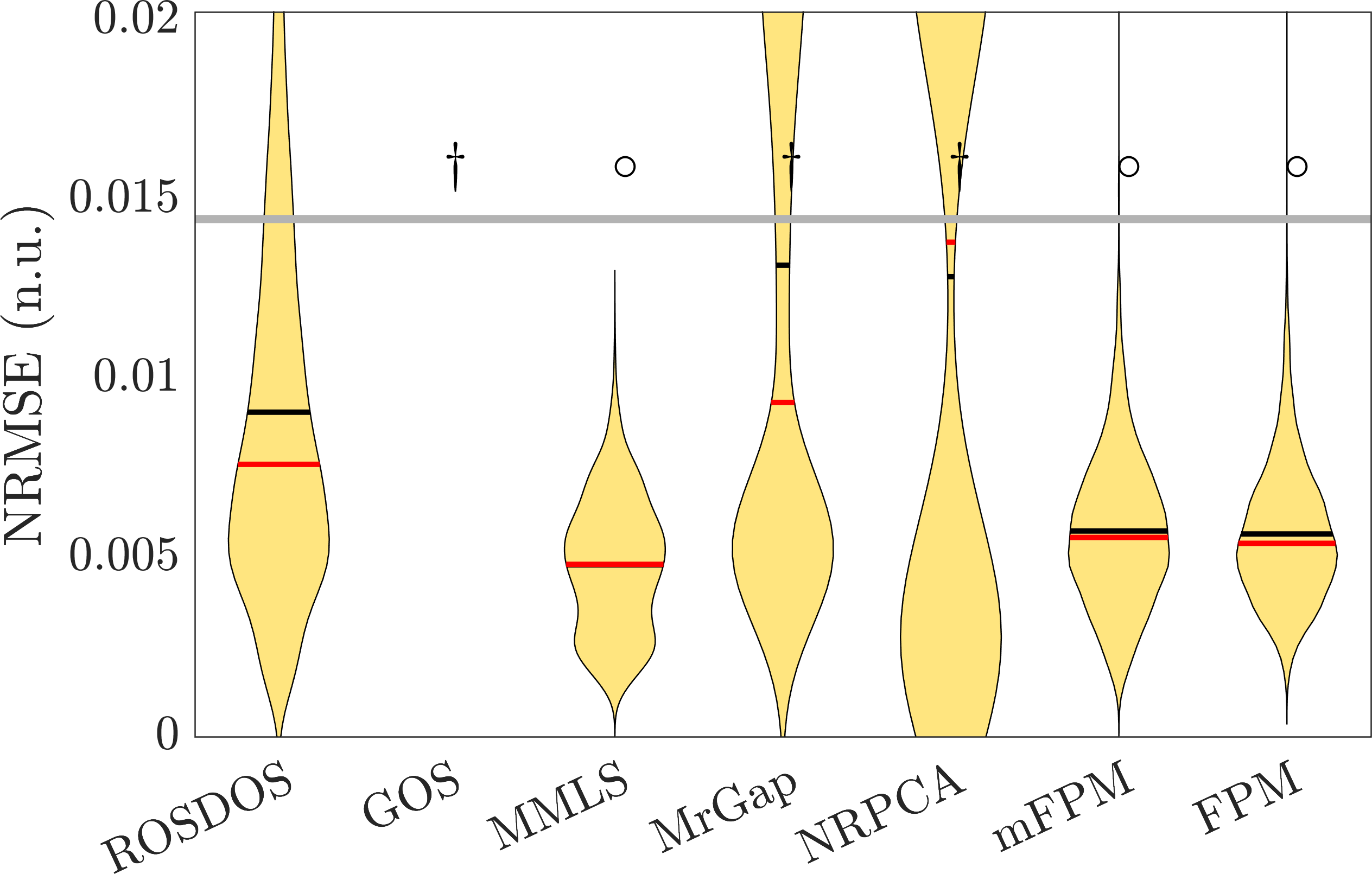} 
        \includegraphics[width=0.32\textwidth]{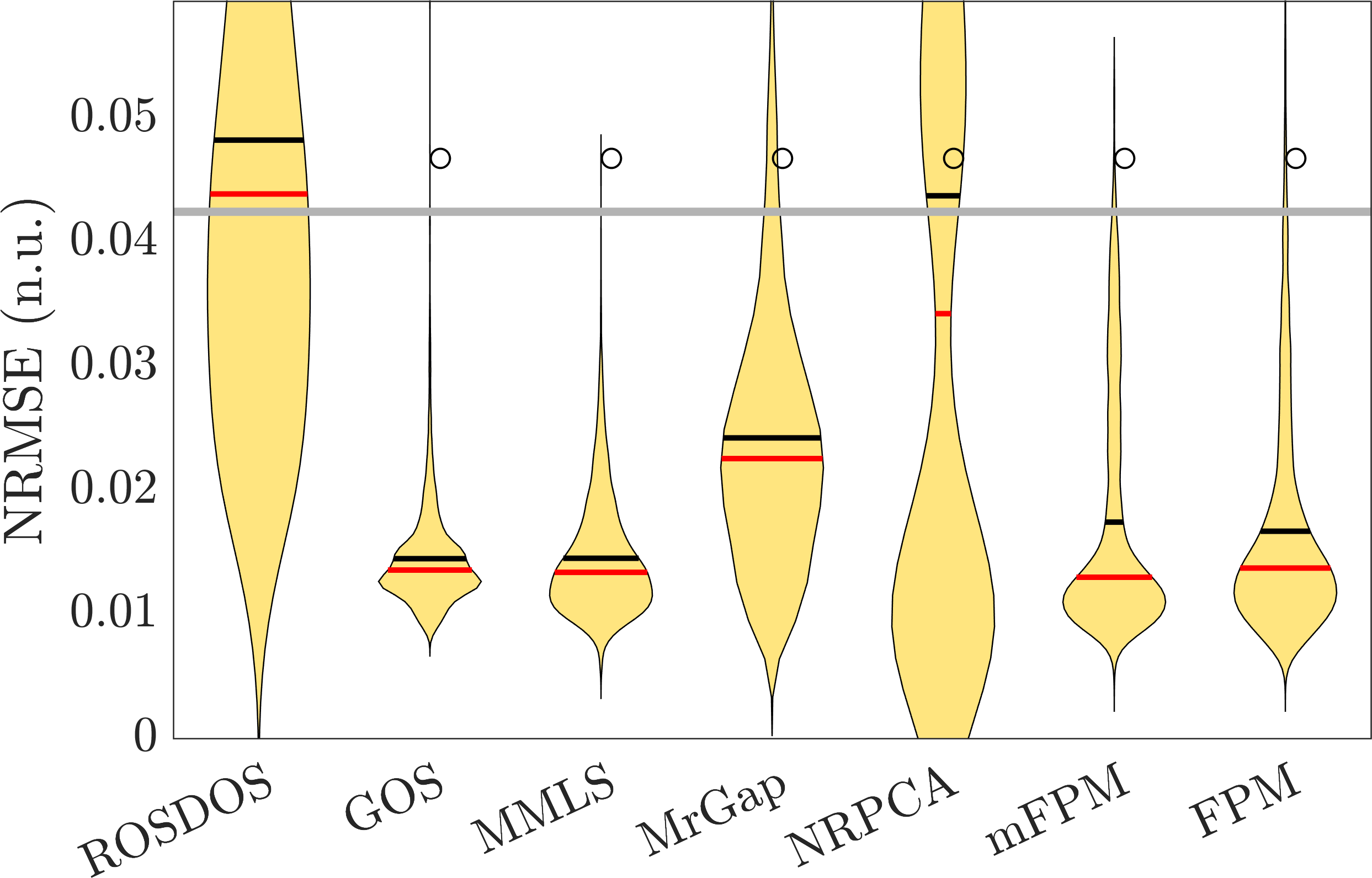} \\
        \includegraphics[width=0.32\textwidth]{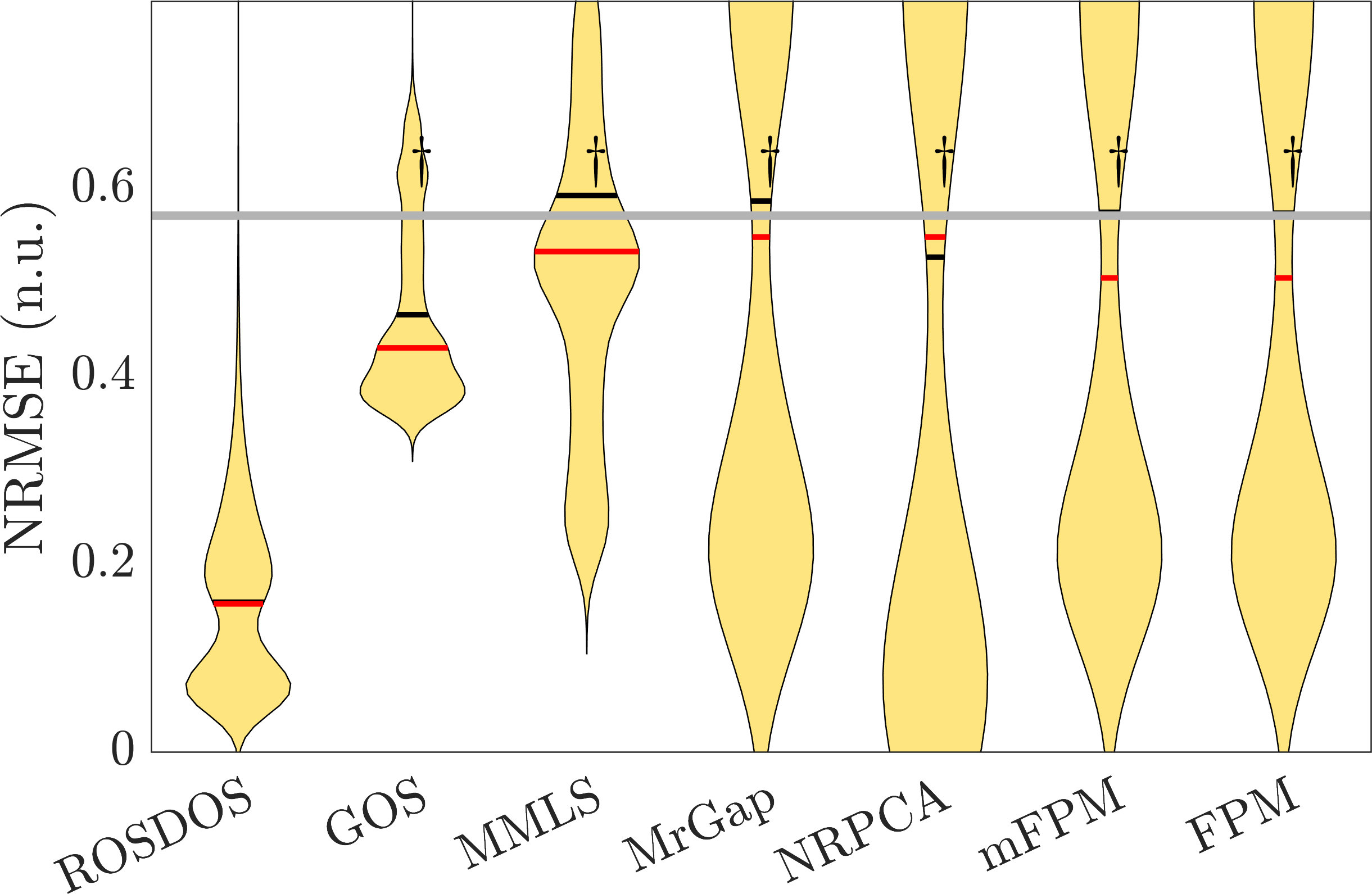} 
        \includegraphics[width=0.32\textwidth]{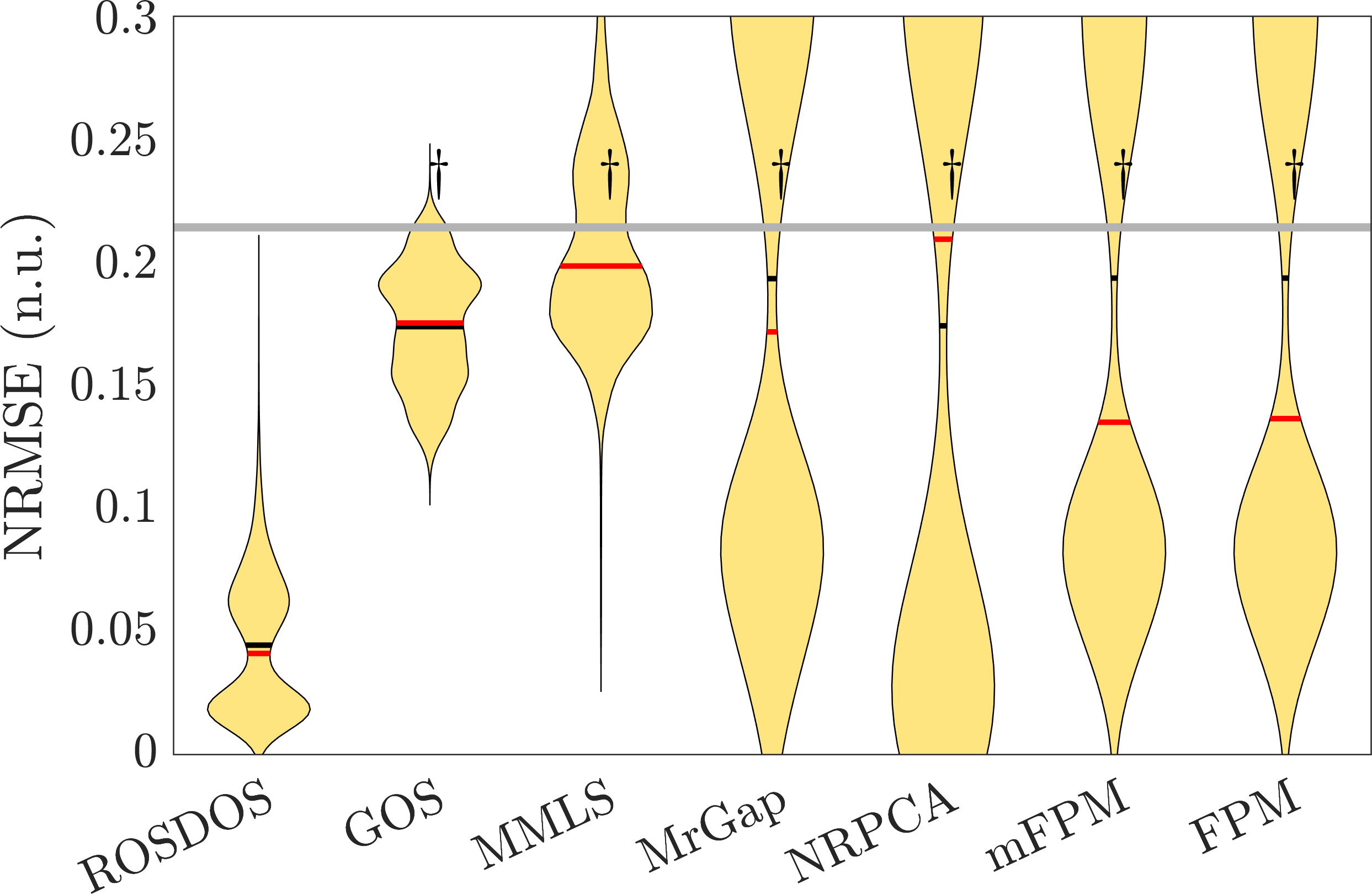} 
        \includegraphics[width=0.32\textwidth]{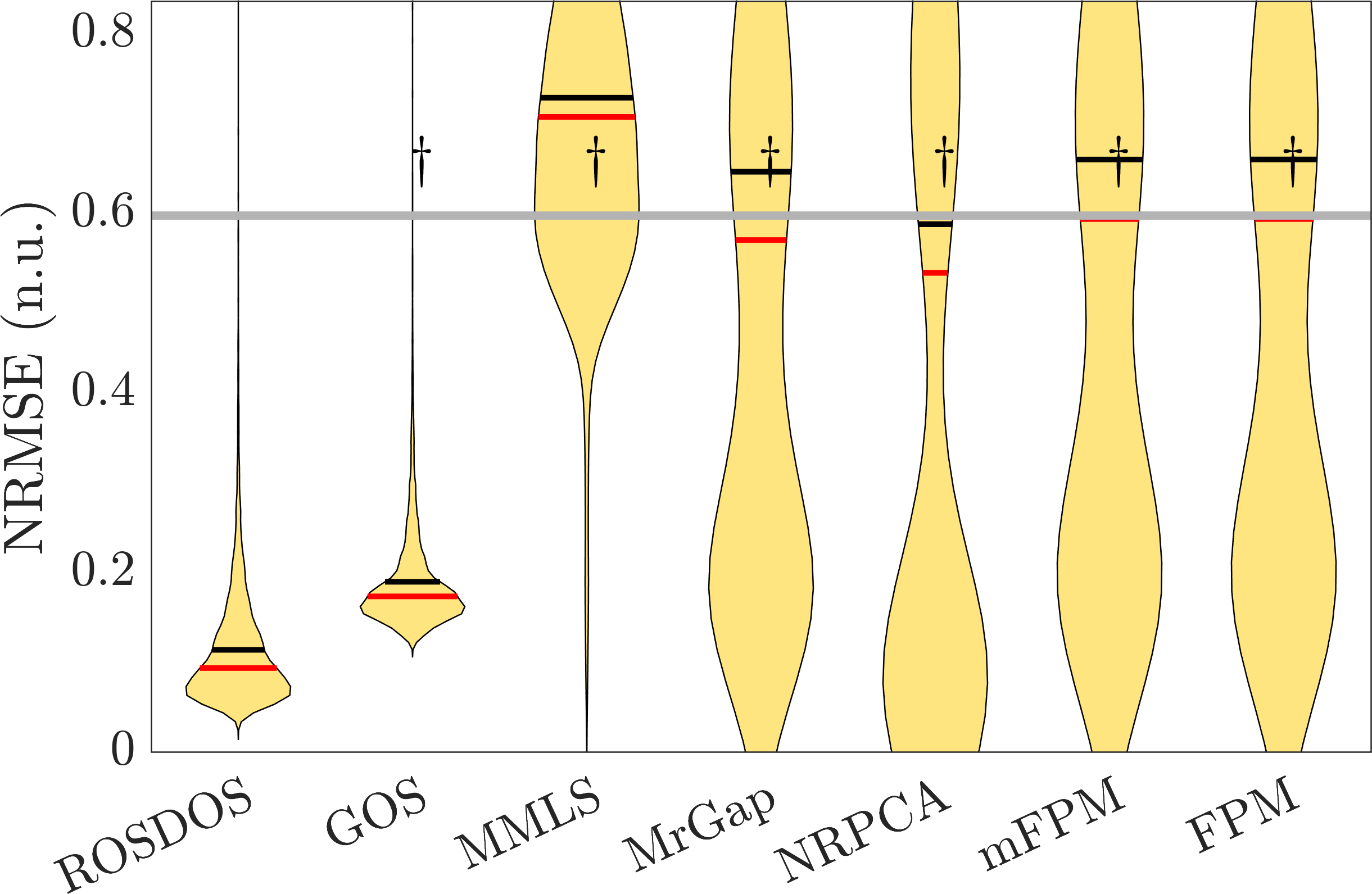} \\
        \includegraphics[width=0.32\textwidth]{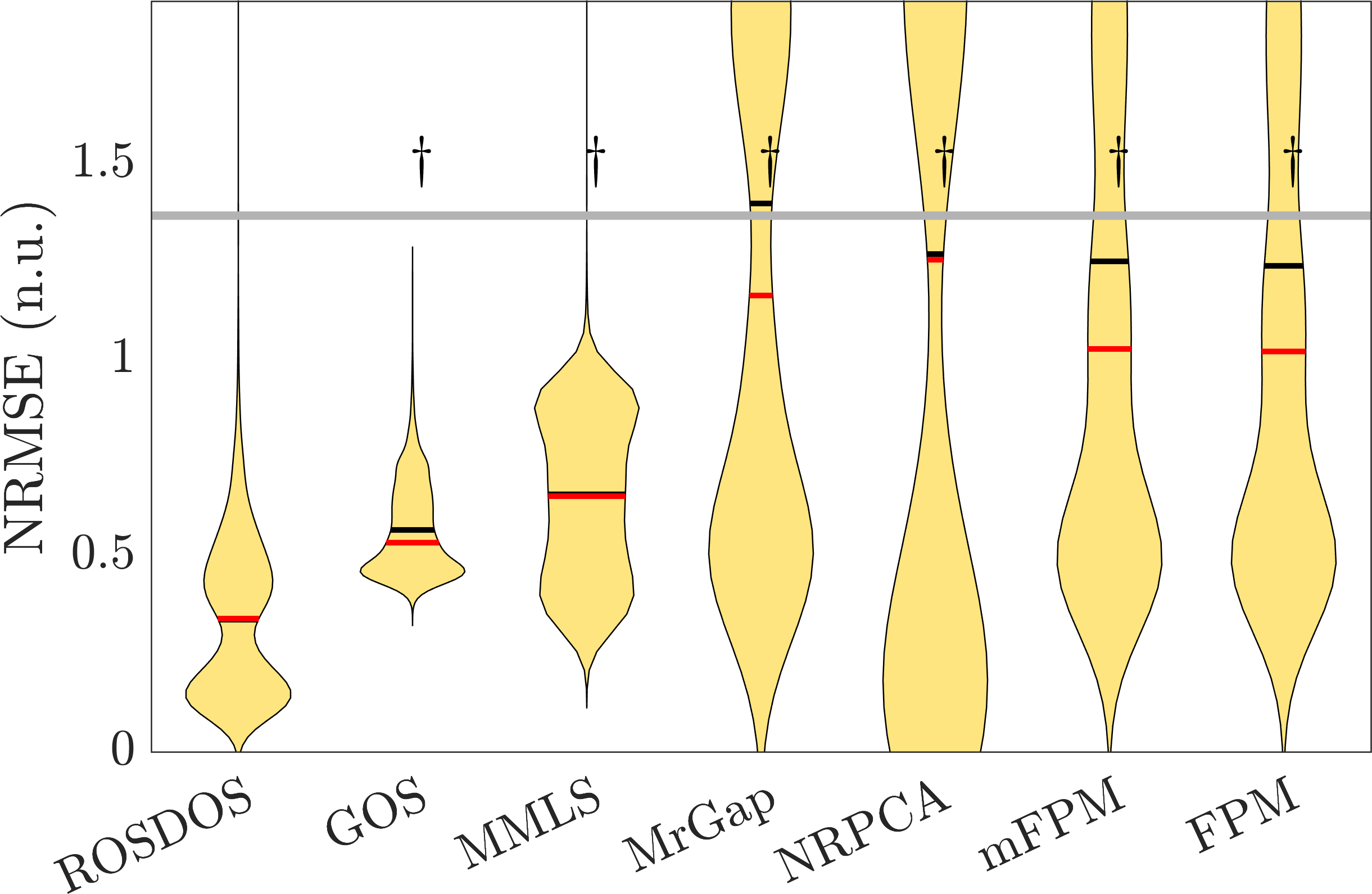}
        \includegraphics[width=0.32\textwidth]{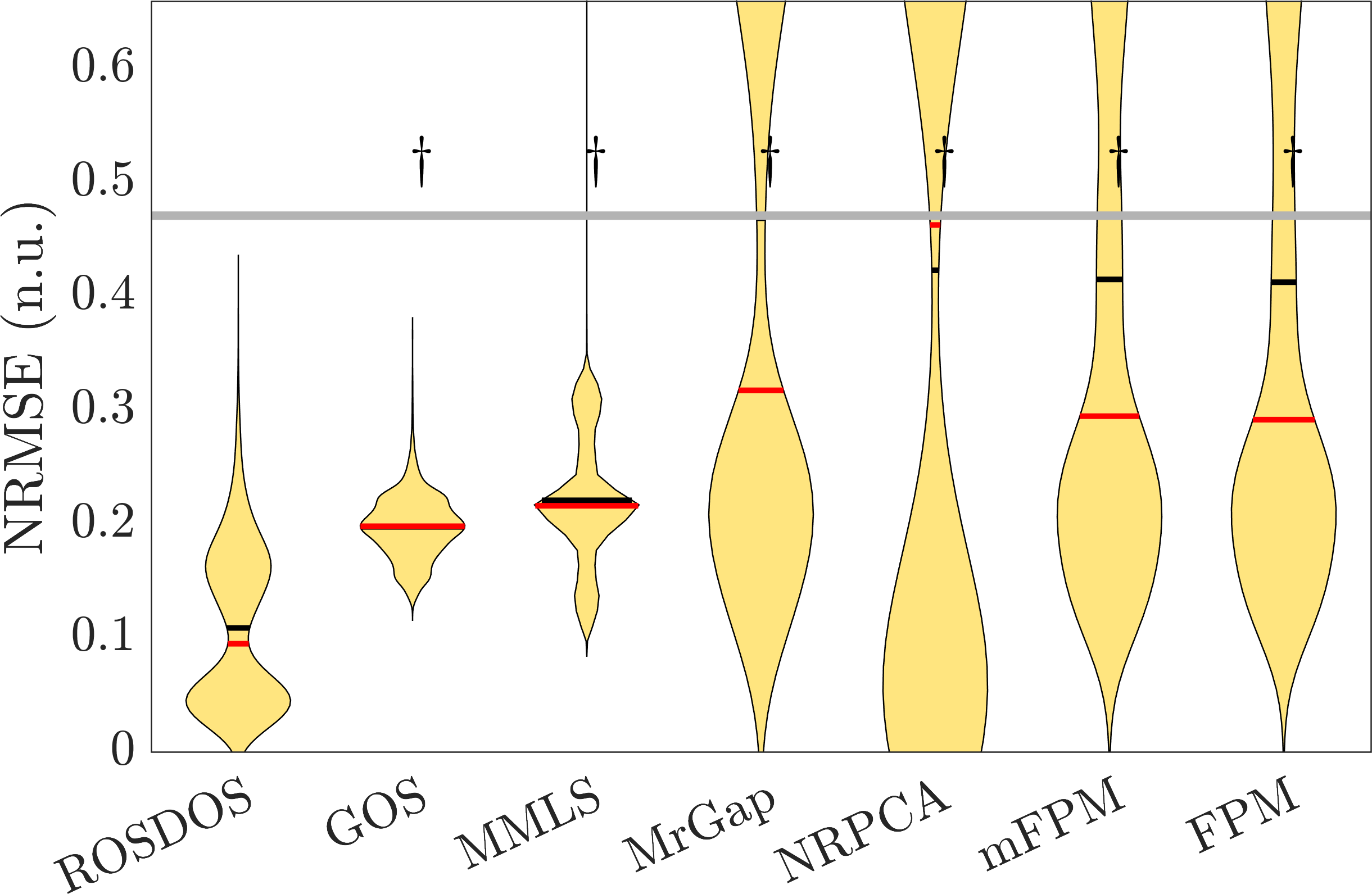} 
        \includegraphics[width=0.32\textwidth]{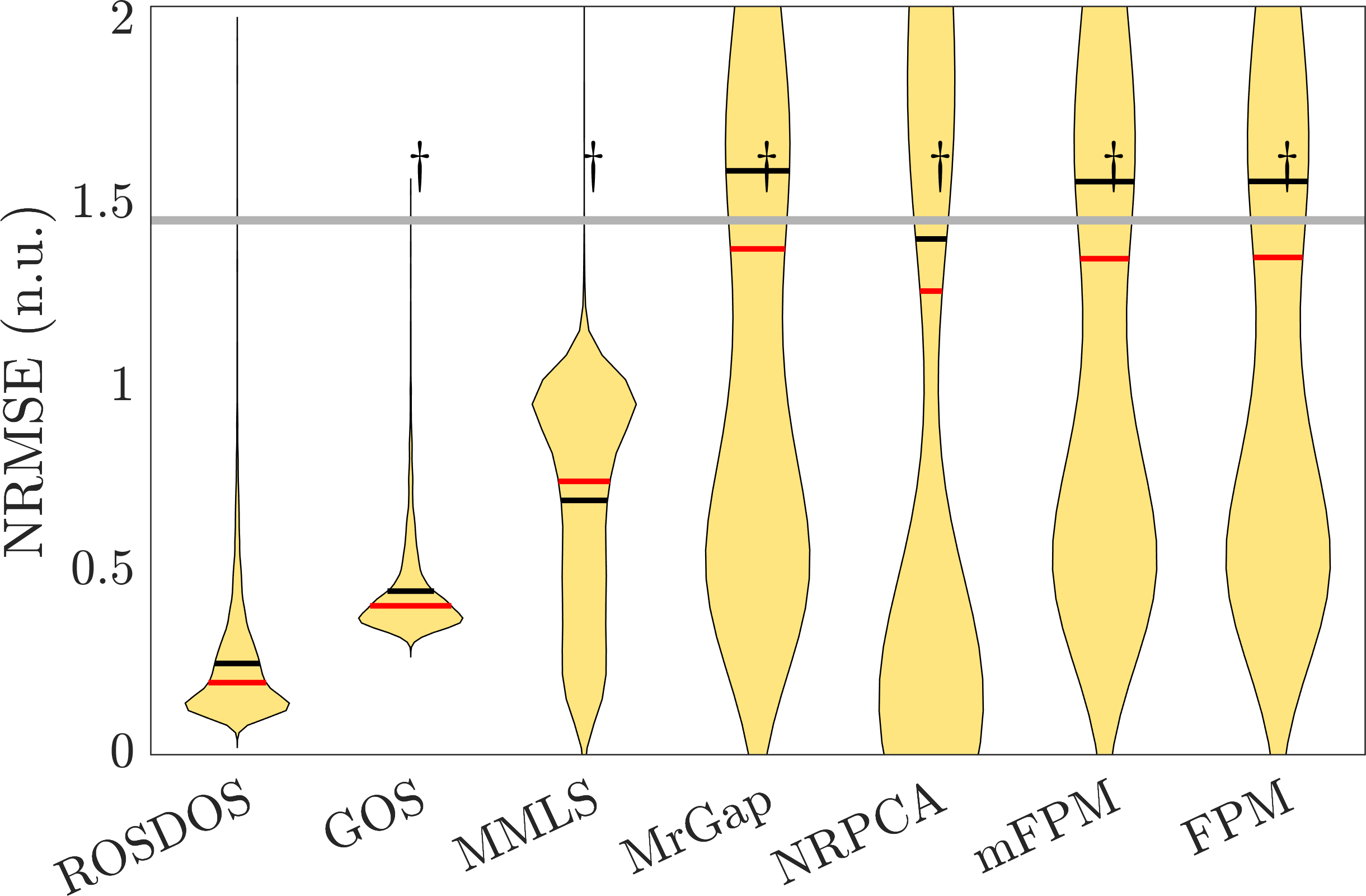} 
            \caption{\label{fig:SimuNRMSE2}A summary of denoising efficiency of different algorithms over $12$ different simulated databases with noise with the separable covariance structure in terms of NRMSE. The distributions of NRMSE of ROSDOS, GOS, MMLS, MrGap, NRPCA, mFPM and FPM are shown in yellow. From left to right columns, the manifolds are $M_1$, $M_2$ and $M_3$. From the first to third rows are associated with $\alpha=1, 1/2$ and $1/3$. For each method, the distribution of NRMSE is estimated using the kernel density estimation with the Gaussian kernel with the optimal kernel bandwidth, which is shown as the violin plot. The gray horizontal line is the median of $\{\|\xi_i\|_2/\|s_i\|_2\}$. The red bar indicates the median and the black bar indicates the mean. To enhance the visualization, the y-axis upper bound is set to $1.4$ times of the median of $\{\|\xi_i\|_2/\|s_i\|_2\}$. The dagger (circle respectively) indicates that \textsf{ROSDOS} performs better (worse respectively) than other manifold denoiser.}
\end{figure}

\begin{figure}[!hbt]
    \centering
 	\includegraphics[trim=0 100 0 0, clip, width=\textwidth]{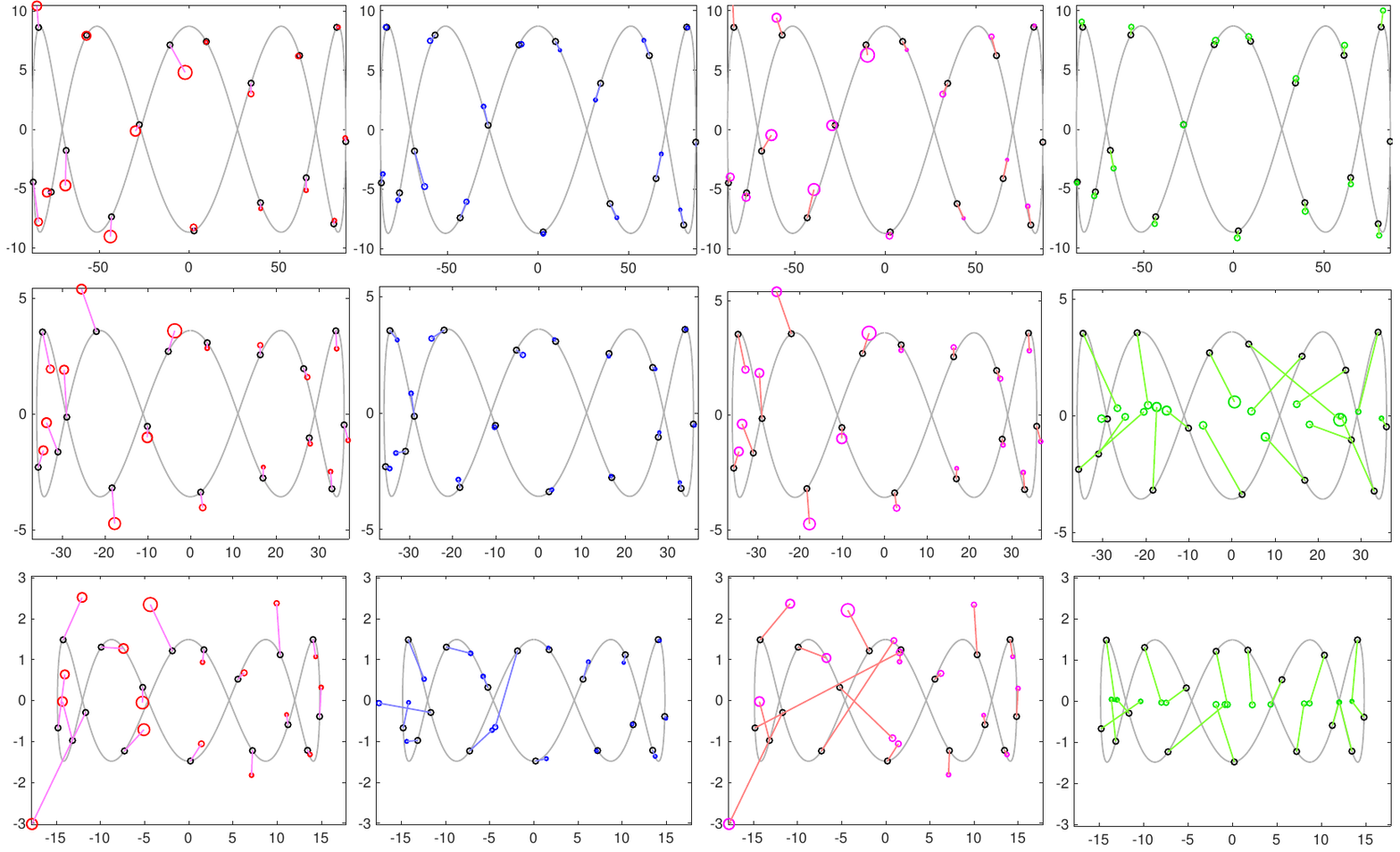} 
            \caption{\label{fig:SimuVisualization}A visualization of different manifold denoisers using $M_1$ contaminated by noise with separable covariance structure as an example, where we show the first and tenth axes of the high dimensional dataset. From left to right columns: the noisy data (red circle), the \textsf{ROSDOS} result (blue circle), the mFPM result (magenta circle) and the MMLS result (green circle) respectively. From top to bottom rows: the mSNR is 26.48 dB, 3.5dB and -4.2dB respectively. The circle size indicates the root mean squared error.}
\end{figure}

\begin{figure}[!hbt]
    \centering
        \includegraphics[width=.7\textwidth]{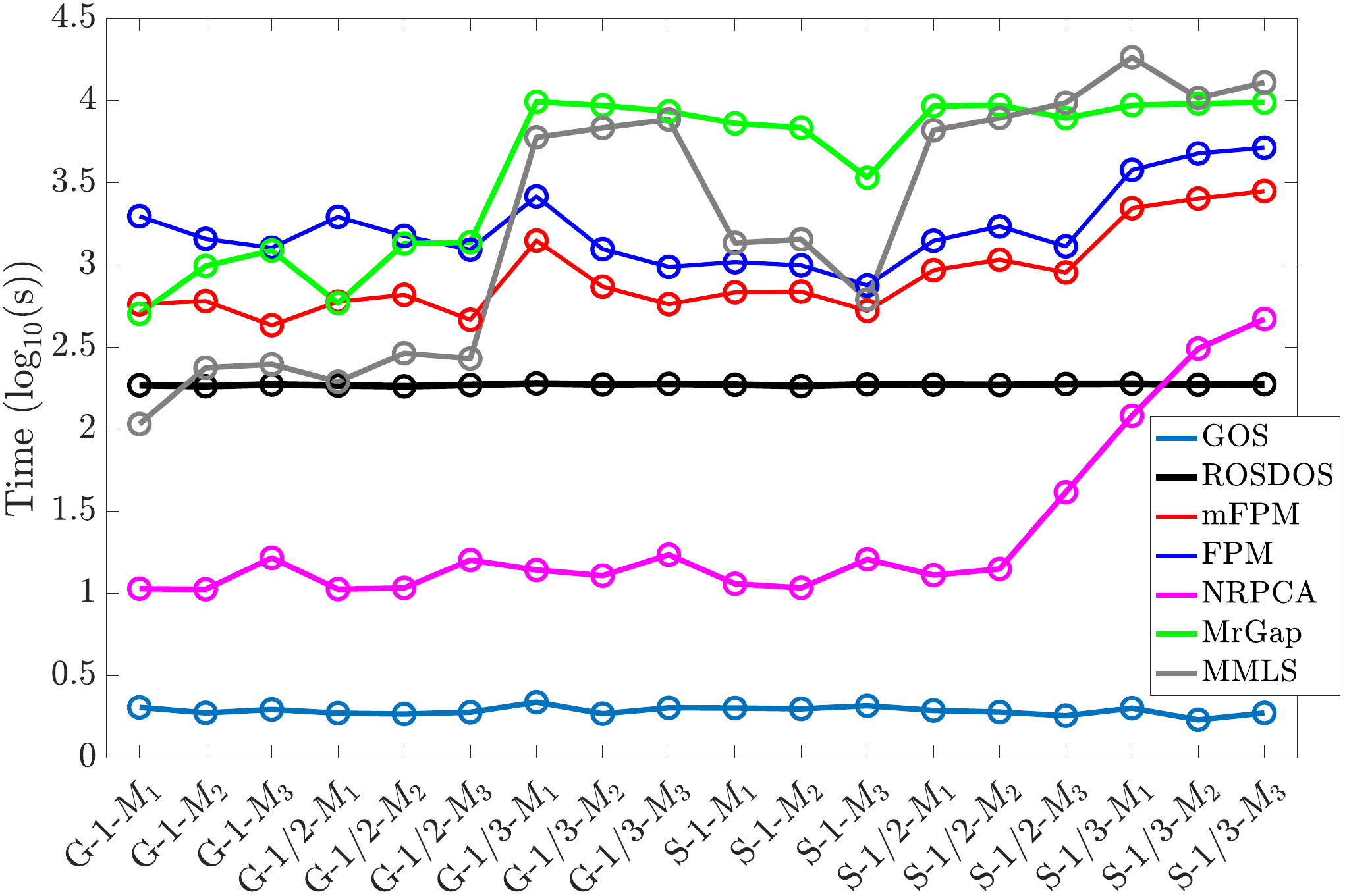} 
    \caption{\label{fig:SimuCompTime}A summary of computational time of different algorithms over $18$ different simulated databases. In the x-axis, G means Gaussian noise, S in the beginning means noise with the separate covariance structure, $1$, $1/2$ and $1/3$ in the middle means $\alpha$, and M1, M2 and M3 in the end means three simulated manifolds. The time is shown in the $\log_{10}$ scale with the unit second.}
\end{figure}

When $\alpha=1$ ($\alpha=1/2$ and $\alpha=1/3$ respectively), the mSNRs are about 30.25 dB (7.2 dB and -0.4dB respectively) for three manifolds when the noise is Gaussian, and 26.48 dB (3.5 dB and -4.2 dB respectively) when the noise satisfies the separable covariance structure. The NRMSE and computational time of different algorithms are shown in Figures \ref{fig:SimuNRMSE} and \ref{fig:SimuCompTime}. 
When noise is negligible ($\alpha=1$), all proposed manifold denoising algorithms perform well, and outperform \textsf{ROSDOS}. In this noise level, the overall noise energy decays to zero at the rate $p^{-1}$ when $p\to \infty$ and the neighbors can be relatively accurately determined , so most algorithms work well with their theoretical supports. Since \textsf{ROSDOS} simply calculate the median of all neighbors, the estimation is biased by curvature. This suggests the potential of combining \textsf{ROSDOS} with other locally nonlinear fitting to further enhance the algorithm when the noise is negligible. 
When noise is moderate ($\alpha=1/2$), \textsf{ROSDOS} starts to outperform other methods, except MMLS when the noise level is Gaussian. In this region, the overall noise energy is asymptotically a positive constant when $p\to \infty$, and the noise impact does not disappear and detailed neighboring information is not accurate, which downgrades the performance of most algorithms. In MMLS, this neighboring information can be ``saved'' by iteratively updating the neighbors and tangent space estimation. 
When noise is ample ($\alpha=1/3$), \textsf{ROSDOS} outperforms other methods. 
In this region, the overall noise energy blows up at the rate $p^{1/3}$, and the neighboring information is no longer trustable, and fully dominated by the noise. The benefit of a robust metric design like \textsf{ROSDOS} could be seen in the simulation.
Also notice that GOS in general performs quite well in $M_3$, since its embedding dimension is fixed to $4$ and hence the low rank property holds. But when the noise level is ample, a non negligible portion of noise remains since GOS is essentially a global hard threshold denoise method like TSVD in our setup.   
To further understand the behavior of different algorithms, we show the partial noisy and denoised datasets in Figure \ref{fig:SimuVisualization}, where we show the first and tenth axes of the high dimensional dataset with $M_1$ contaminated by noise with the separable covariance structure. When noise is small, both mFPM and MMLS perform well as shown in Figure \ref{fig:SimuNRMSE2}. When noise becomes large, \textsf{ROSDOS} survives and can successfully remove several large noises. 

In terms of computational efficiency, both NRPCA and GOS exhibit significantly lower computational times when compared to alternative algorithms. On the contrary, MMLS and MrGap necessitate a comparatively higher amount of computational time, particularly when the noise has a separable covariance structure. It is worth noting that the computational efficiency of \textsf{ROSDOS} differs from iterative algorithms. In the case of \textsf{ROSDOS}, its computational time predominantly depends on $p$ and $n$. Consequently, the computational time remains consistent across simulated datasets, reflecting a distinctive characteristic of \textsf{ROSDOS} as a non-iterative algorithm.

\subsection{Semi-realistic LFP during DBS}\label{section: LFP DBS example}

While there are many other examples like stimulus artifacts in the intracranial EEG \cite{alagapan2019diffusion} and the trans-abdominal maternal ECG \cite{su2019OSonEEG}, we focus on the LFP in this section.  
Deep brain stimulation (DBS) has been proved to be an effective treatment for various diseases, like Parkinson's disease if it is the high-frequency stimulation (HFS) in the subthalamic nucleus \cite{limousin1995DBS}. Through the implementation of a pair of electrodes, neural activity can be accurately recorded via externalization of leads as LFPs, which offer valuable insights into our understanding of brain dynamics associated with different diseases, with distinct frequency ranges corresponding to different clinical symptoms \cite{neumann2019aDBSreview}. However, during HFS, stimulus artifacts are inevitably recorded along with LFPs, which contaminates the LFP in the form of high frequency oscillation. Thus, decomposing the stimulation artifacts from the recorded LFP is a critical mission toward exploring LFP and hence the brain dynamics during DBS. Several algorithms have been proposed in the literature, which can be roughly classified into the filtering approach, the blanking approach and the template subtraction approach. See \cite{LFPOS2023} for a summary of these methods. However, due to the nonstationary and broad spectrum nature of the stimulation artifacts, these methods are all limited in recovering LFP during HFS. 

\begin{figure}[!hbt]
    \centering
    \begin{tabular}{cc}
        \includegraphics[width=0.3\textwidth]{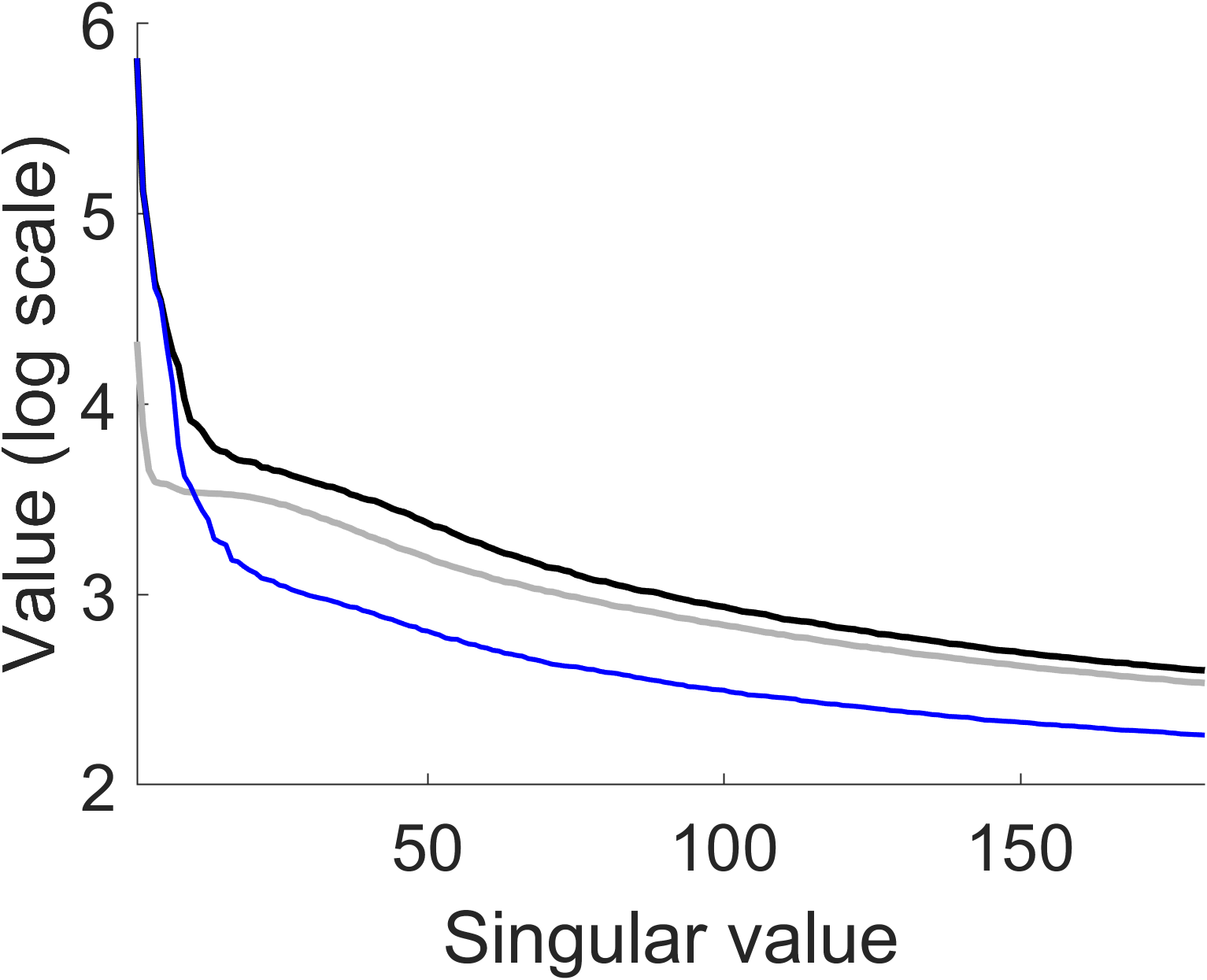} 
        & \includegraphics[trim=0 0 0 0, width=0.65\textwidth]{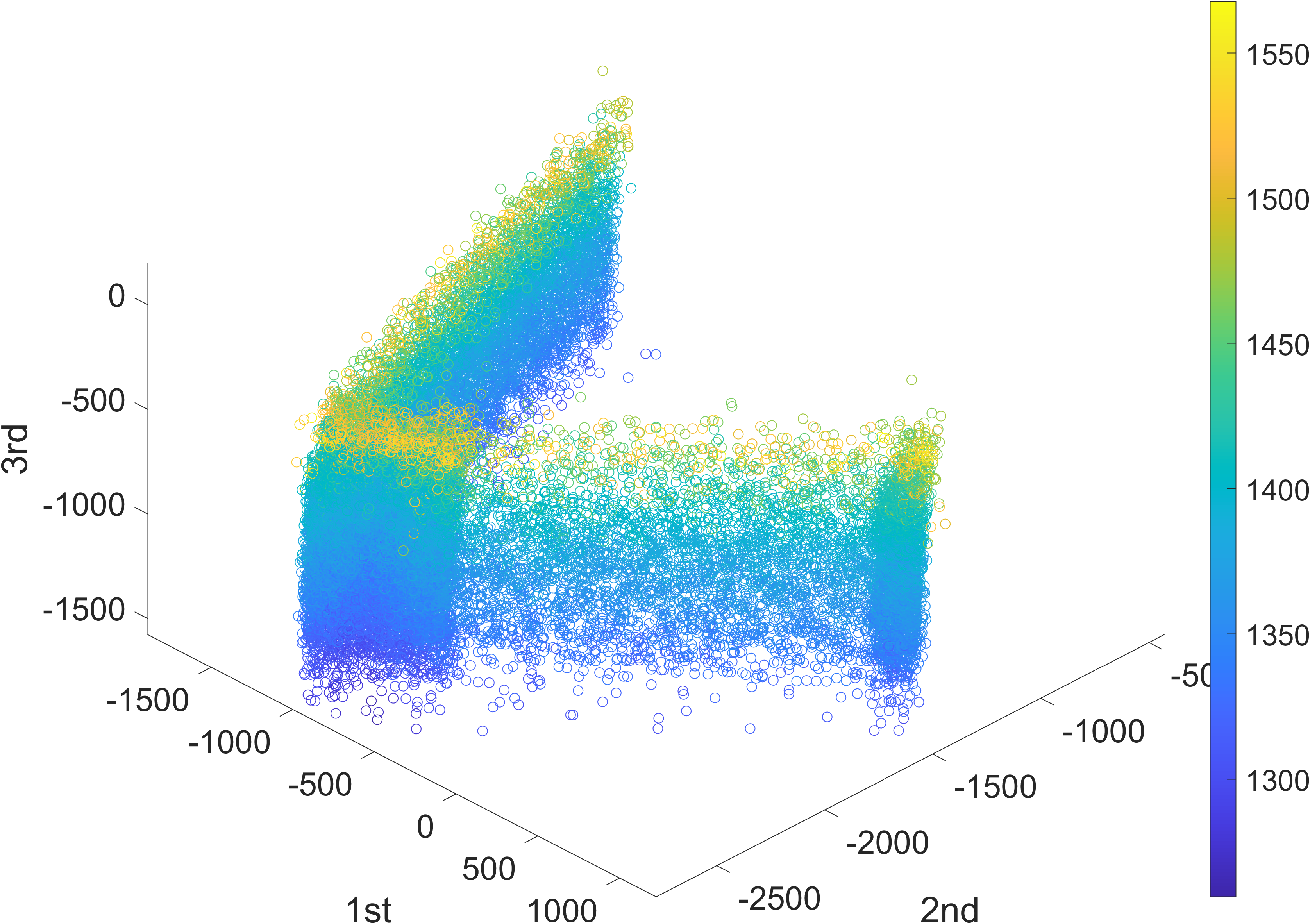}\\
        \scriptsize  (a) & \scriptsize (b)
    \end{tabular}
    \caption{\label{fig:SVDandPCA} An illustration of the nonlinear structure of the stimulus artifacts. (a) The black line shows the singular values of the data matrix $\SX$ in the log scale. (b) The dimension of $\SX$ is reduced to three by PCA, and the color represents the amplitude of each stimulus artifact.}
\end{figure}

To resolve the LFP challenge, an algorithm called {\em Shrinkage and Manifold based Artifact Removal using Template Adaptation} (SMARTA) was proposed in \cite{LFPOS2023}. SMARTA is a special version of the proposed manifold denoiser \textsf{ROSDOS}, which is classified as the template subtraction approach. The basic idea behind the template subtraction is that all stimulation artifacts look ``similar'', so if we could group them properly, we could recover the stimulation artifacts by averaging them to remove the LFP based on the law of large number, and hence the LFP could be obtained by subtracting the recovered stimulation artifacts from the recorded LFP. To achieve it, the recorded LFP is converted in the following way. 
After preprocessing the LFP signal recorded during DBS at 44k Hz, including removing the low-frequency noise by a high-pass filter with a cut-off frequency of 3Hz and determining the location of each artifact if the stimulation artifacts' locations are unknown, the signal is divided into segments so that each segment begins 1.2ms before and ends 7ms after the corresponding peak location and contains a single stimulus artifact. Suppose we obtain $n$ segments ${x}_1,\, {x}_2,\ldots,{x}_n\in \mathbb{R}^p$ and hence a data matrix 
\begin{align}\label{definition XX}
\XX=\begin{bmatrix}{x}_1 & {x}_2 & \ldots & {x}_n \end{bmatrix}=\SX+\boldsymbol\Xi\in \mathbb{R}^{p\times n}\,,
\end{align}
where, $p$ is the length of a segment, $\SX$ contains pure stimulation artifacts, and $\boldsymbol\Xi$ contains ``clean'' LFP. Thus, all truncated stimulation artifacts are aligned. In an {\em ideal} situation that all stimulus artifacts are the same up to a scaling, $\SX$ is of rank $1$; that is $\SX=\sigma\boldsymbol u\boldsymbol v^\top$, where $\sigma>0$, $\boldsymbol u\in O(p)$ is the template for the stimulus artifact and $\boldsymbol v\in O(n)$ quantifies the scaling of each stimulus artifact; that is, $\boldsymbol v(i)$ is the scale of the $i$-th stimulus artifact. However, in practice, due to the complicated nonlinear interaction between electrical stimulus and the brain, the stimulus artifacts are different from one to another. This complicated nonlinear structure is modeled by
\begin{align}\label{SX model clean}
\SX=\sum_{l=1}^r \sigma_l\boldsymbol u_l\boldsymbol v_l^\top=\begin{bmatrix} {{s}}_1 & {{s}}_2 & \ldots & {{s}}_n \end{bmatrix}, 
\end{align}
where $\sigma_1\geq \sigma_2\geq \ldots \geq \sigma_r$, $r\leq \min\{n,p\}$ might {\em not} be low. 
Moreover, the stimulus artifacts are in general not random but follow some nonlinear structures. We thus model these stimulus artifacts to be parametrized by a $d$ dimensional manifold $M$ that is isometrically embedded in an $r$-dim affine subspace of $\mathbb{R}^p$. To demonstration this model, take a semi-real LFP during DBS from \url{https://doi.org/10.7910/DVN/BN1PIR}, and construct a data matrix $\SX$ as in \eqref{definition XX} with the size $n=25,836$ and $p=181$. 
In Fig.~\ref{fig:SVDandPCA}(a), the singular values of $\SX$ are shown as the black curve. While the largest and smallest singular values are different by about 3 orders, the tail singular values of $\SX$ are still significant. This suggests that $r$ is in general high.
To observe the low dimensional nonlinear structure of the stimulus artifacts, the principal component analysis (PCA), is applied. The three dimensional embedding of $\SX$ by projecting $\SX$ to the top three principal vectors is shown in Fig.~\ref{fig:SVDandPCA}(b), where the color labels the amplitude for each stimulus artifact. A clear nonlinear structure can be observed with the stimulus artifact amplitude as one parametrization parameter. Since PCA is linear, the nonlinear structure is preserved, which provides evidence that the stimulus artifacts follow a nonlinear structure,

For the noise matrix that contains ``clean'' LFP
\begin{align}\label{definition ZX}
\boldsymbol\Xi=\begin{bmatrix}{\xi}_1 & {\xi}_2 & \ldots & {\xi}_n \end{bmatrix}\in \mathbb{R}^{p\times n},
\end{align}
clearly it inherits the long range dependent structure of LFPs \cite{bedard2006does}. Thus the noise matrix cannot be white and independent. We choose to model $\boldsymbol\Xi$ so that it satisfies the separable covariance structure \cite{filipiak2017comparison}; that is, the covariance of $\boldsymbol\Xi$ can be written as $ A\otimes  B$, where $ A\in \mathbb{R}^{p\times p}$ models the color structure of the LFP in each segment and $ B\in \mathbb{R}^{n\times n}$ models the dependence across different segments.

We now compare different methods on the semi-realistic LFP during HFS using an agar brain model
The simulated data matrix is composed of the LFP recorded from patients without stimulation (or resting state), denoted as $\boldsymbol{\Xi}\in \mathbb{R}^{p\times n}$, and the stimulus artifacts at 130Hz generated from the agar brain model, denoted as $\boldsymbol S\in \mathbb{R}^{p\times n}$, where $p=181$ and $n=5000$. The simulated dataset is available at \url{https://doi.org/10.7910/DVN/BN1PIR}. See \cite[Section 2.2]{LFPOS2023} for details.
Clearly, the LFP as noise is not white. The noise matrix $\boldsymbol\Xi$ is first scaled so that the determinant of the covariance of normalized noise matrix, also denoted as $\boldsymbol\Xi$, is $1$. The clean matrix is scaled by the same factor, also denoted as $\boldsymbol S$. Then, we consider three setups with different SNRs, $\boldsymbol{X}^{(s)}=s\boldsymbol{S}+\boldsymbol\Xi$, where $s\in\{2,1/3,1/14\}$. The mSNRs of $ \boldsymbol{X}^{(2)}$, $\boldsymbol{X}^{(1/3)}$ and $\boldsymbol{X}^{(1/14)}$ are 31.30, 15.74 and 2.36 respectively.

\begin{figure}[!hbt]
    \centering
        \includegraphics[trim=0 0 0 0, clip, width=\textwidth]{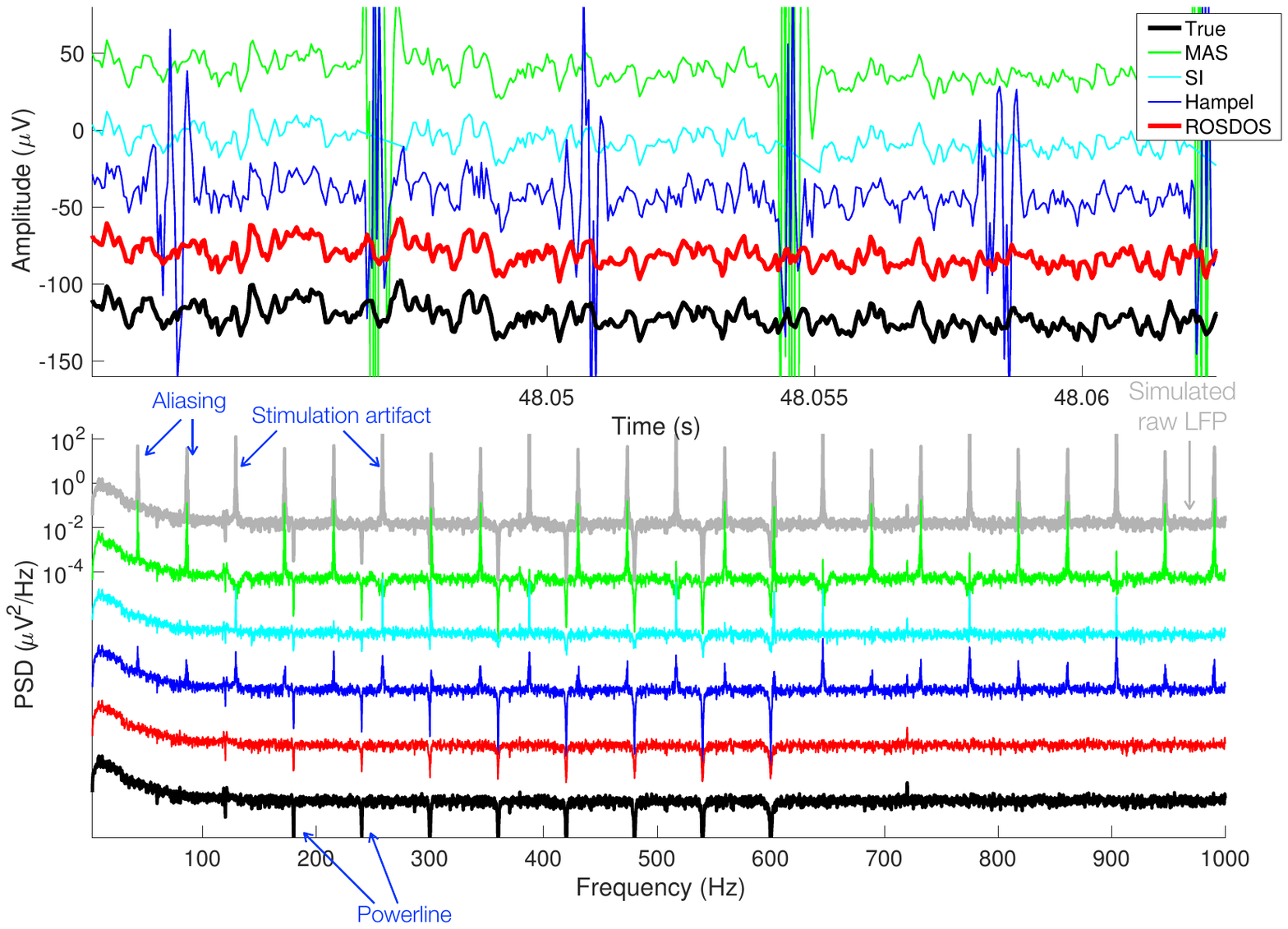} 
    \caption{\label{fig time freq domain DBS}
    An illustration of stimulation artifact removal efficiency of different existing algorithms over the semi-real LFP database during DBS shown in the time domain (top) and frequency domain (bottom). To enhance the visibility, the results of different algorithms in both time and frequency domains are shifted vertically. The power spectrum of the raw data is shown in the gray curve. The artifacts introduced by aliasing, stimulation and power line are indicated by blue arrows.}
\end{figure}

\begin{figure}[!hbt]
    \centering
    \begin{minipage}{0.5\textwidth}
        \includegraphics[width=0.9\textwidth]{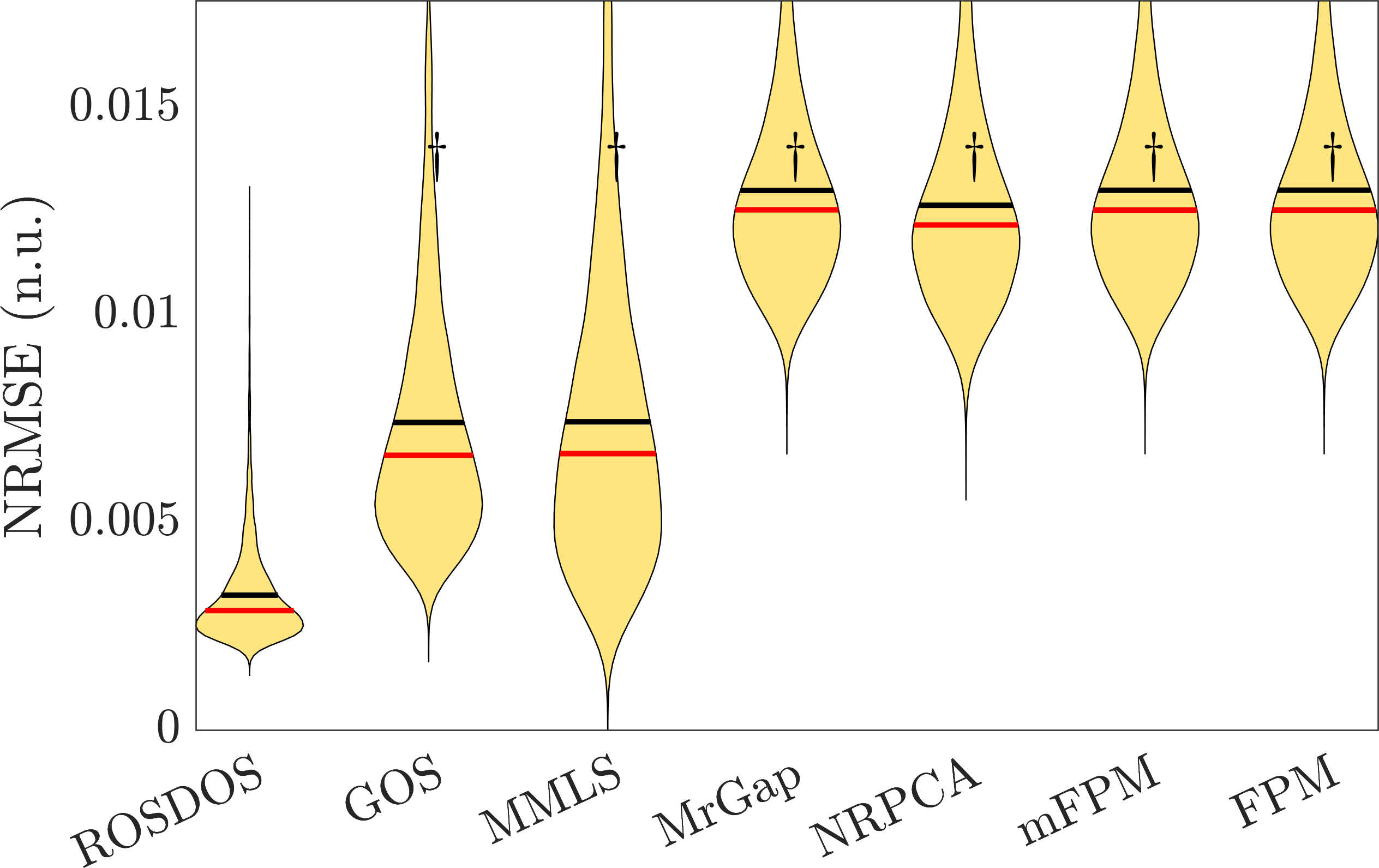} \\
        \includegraphics[width=0.9\textwidth]{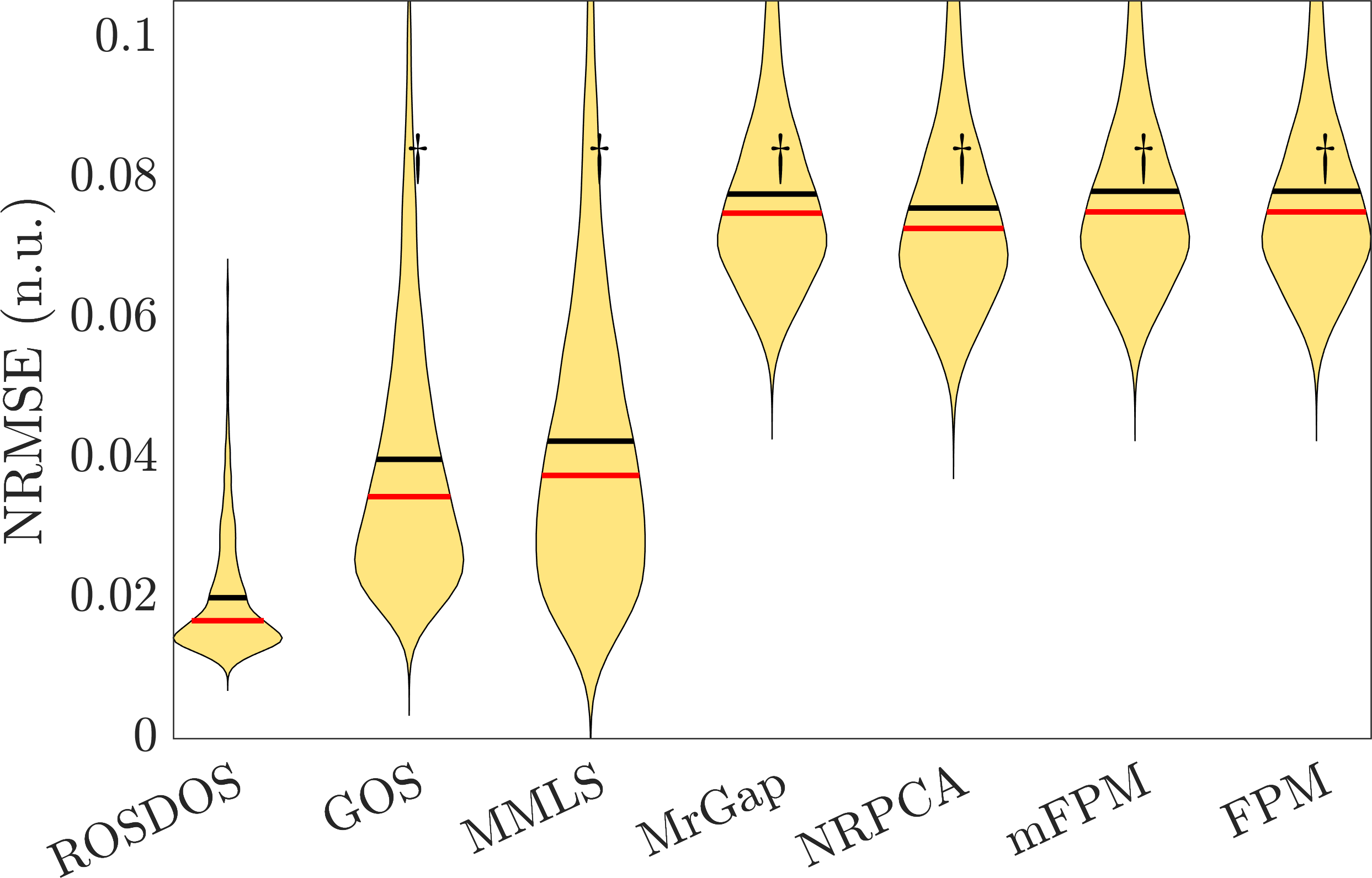} \\
        \includegraphics[width=0.9\textwidth]{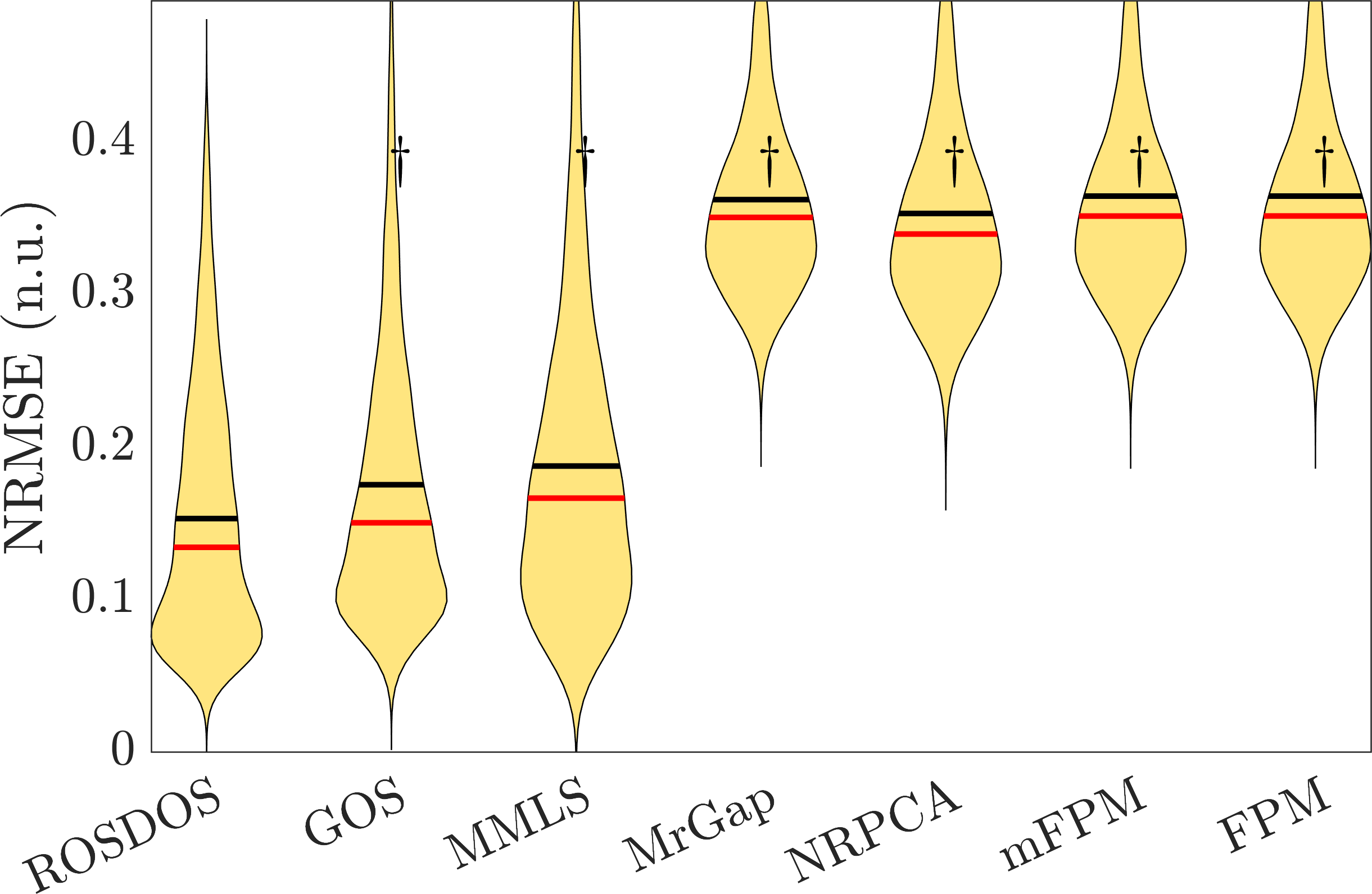} 
        \end{minipage}
        \begin{minipage}{0.49\textwidth}        
         \includegraphics[width=0.8\textwidth]{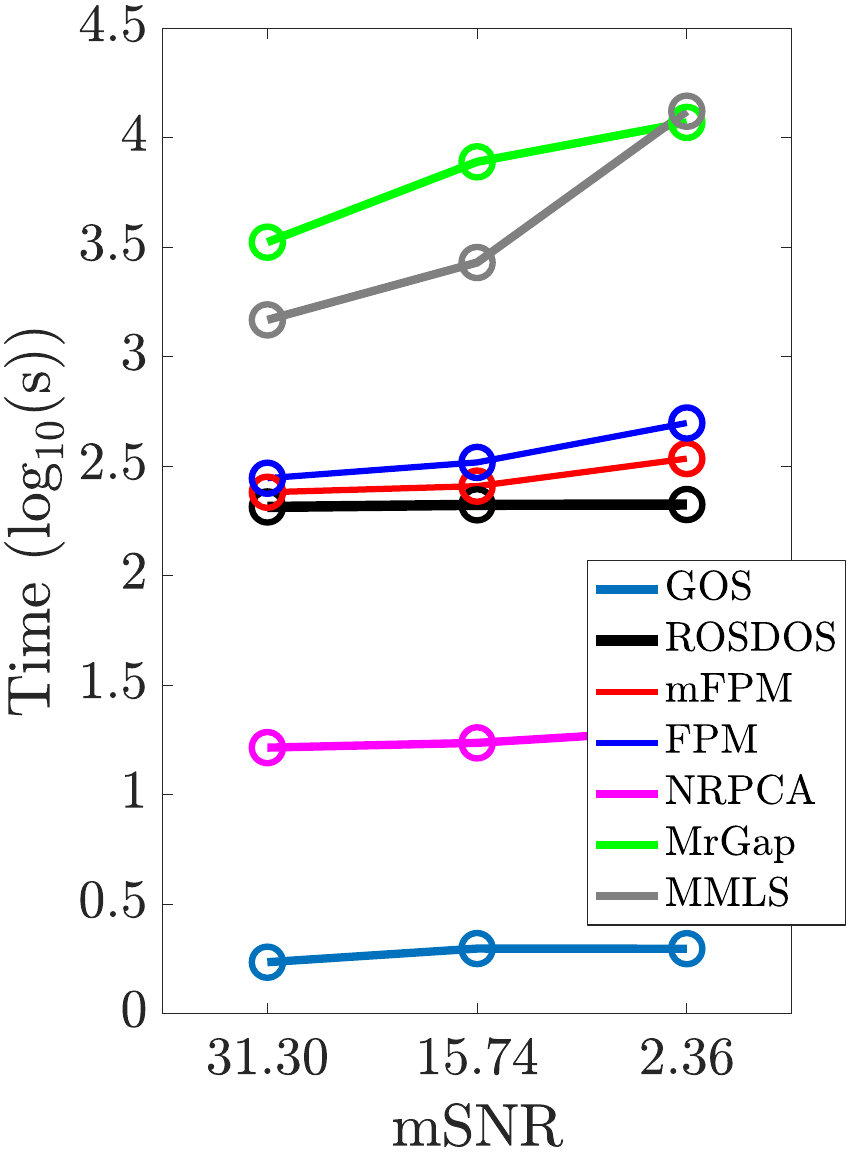} 
         \end{minipage}
    \caption{\label{fig result DBS}
    A summary of denoising efficiency of different algorithms over the semi-real LFP database during DBS in terms of NRMSE and their computational times. The distributions of NRMSE of ROSDOS, GOS, MMLS, MrGap, NRPCA, mFPM and FPM are shown in yellow. From left top to right bottom, $s=2,1/3,1/14$. For each method, the distribution of NRMSE is estimated using the kernel density estimation with the Gaussian kernel with the optimal kernel bandwidth, which is shown as the violin plot. 
    The red bar indicates the median and the black bar indicates the mean. To enhance the visualization, the y-axis upper bound is set to $1.4$ times of the median of $\{\|\xi_i\|_2/\|s_i\|_2\}$. The dagger (circle respectively) indicates that \textsf{ROSDOS} performs better (worse respectively) than other manifold denoiser. On the right hand side, the computational time of different manifold denoising algorithms are shown.}
\end{figure}

A comparison of the application of \textsf{ROSDOS} with existing stimulation artifact removal algorithms, including moving average subtraction (MAS) \cite{sun2016MVS}, sample-and-interpolate (SI) \cite{heffer2008interpolation} and Hampel filtering (HF) \cite{allen2010hampelFiltering}, in both time and frequency domains are shown in Figure \ref{fig time freq domain DBS}. Compared with other algorithms, it is clear that in the time domain, the stimulation artifacts have been well removed, which is reflected in the frequency domain, particularly in the high frequency region (about 200 Hz).
The clean data matrix $\mathbb{S}$ is of full rank, and the effective rank is determined to be $9$. The dimension of the manifold is estimated to be $5$. The summary of performance of different manifold denosing algorithms applied to remove stimulation artifacts in terms of NRMSE is shown in Figure \ref{fig result DBS}. We could see that \textsf{ROSDOS} overall performs significantly better than other algorithms. More complete comparisons will be reported in official paper of this manuscript.

\bibliographystyle{plain}
\bibliography{mybibfile}

\end{document}